\documentclass[11pt]{article}

\usepackage{acl}

\usepackage{times}
\usepackage{latexsym}

\usepackage[T1]{fontenc}

\usepackage[utf8]{inputenc}

\usepackage{microtype}

\usepackage{inconsolata}

\usepackage{graphicx}

\usepackage{bm}
\usepackage{amssymb}
\usepackage{amsmath}
\usepackage{algorithm}
\usepackage{algorithmic}
\usepackage{subcaption}
\usepackage{tabularx}
\usepackage{booktabs}

\usepackage[most]{tcolorbox}
\usepackage{xcolor}

\definecolor{promptgray}{RGB}{245, 245, 245}
\definecolor{headergray}{RGB}{230, 230, 230}

\newtcolorbox{promptbox}[1]{
    colback=promptgray,        
    colframe=black!70,         
    left=5pt, right=5pt, top=5pt, bottom=5pt, 
    arc=2pt,                   
    boxrule=0.5pt,             
    fonttitle=\bfseries\sffamily,
    colbacktitle=headergray,   
    coltitle=black,            
    title=#1,                  
    enhanced,                  
    attach boxed title to top left={yshift=-2mm, xshift=2mm}, 
    boxed title style={boxrule=0.5pt, sharp corners=all}      
}

\usepackage{booktabs}  
\usepackage{multirow}  

\title{Efficient Paths and Dense Rewards: Probabilistic Flow Reasoning for Large Language Models}

\author{Yan Liu$^{1,2}$\thanks{Equal contribution.}\thanks{Work done during internship at Meituan.} \quad Feng Zhang$^{1,3 *}$ \quad Zhanyu Ma$^{1}$ \quad Jun Xu$^{1}$\thanks{Corresponding authors.} \quad Jiuchong Gao$^{1 \ddagger}$ \\ \bf{
Jinghua Hao$^{1}$ \quad Renqing He$^{1}$ \quad Han Liu$^{4}$ \quad Yangdong Deng$^{2}$ }  \\
$^1$Meituan, $^2$Tsinghua University, $^3$Peking University, $^4$Dalian University of Technology
}

\begin{document}
\maketitle

\begin{abstract}

High-quality chain-of-thought has demonstrated strong potential for unlocking the reasoning capabilities of large language models. However, current paradigms typically treat the reasoning process as an indivisible sequence, lacking an intrinsic mechanism to quantify step-wise information gain. This granularity gap manifests in two limitations: inference inefficiency from redundant exploration without explicit guidance, and optimization difficulty due to sparse outcome supervision or costly external verifiers. In this work, we propose \textbf{CoT-Flow}, a framework that reconceptualizes discrete reasoning steps as a continuous probabilistic flow, quantifying the contribution of each step toward the ground-truth answer. Built on this formulation, CoT-Flow enables two complementary methodologies: flow-guided decoding, which employs a greedy flow-based decoding strategy to extract information-efficient reasoning paths, and flow-based reinforcement learning, which constructs a verifier-free dense reward function. Experiments on challenging benchmarks demonstrate that CoT-Flow achieves a superior balance between inference efficiency and reasoning performance.

\end{abstract}

\section{Introduction}

\begin{figure*}[t]
    \centering
    \includegraphics[width=\textwidth]{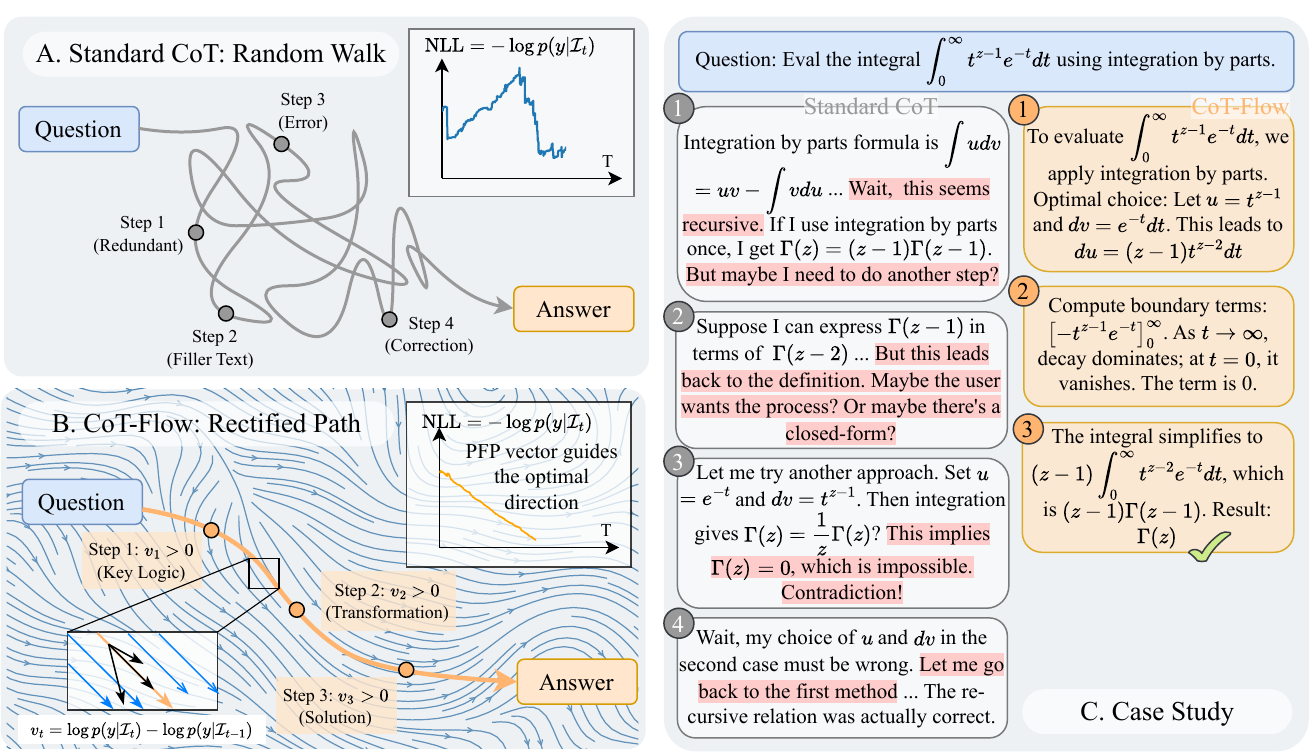}
    
    \caption{Illustration of our proposed CoT-Flow. 
    Panel A: Standard CoT reasoning often exhibits a random walk behavior in the information space, characterized by unstructured exploration, redundancy, and reliance on sparse outcome-based rewards. 
    Panel B (Ours): CoT-Flow models reasoning as a probabilistic flow. By optimizing probabilistic flow progress, it rectifies the reasoning trajectory into the shortest path from the question to the answer, providing dense supervision signals without external verifiers.}
    \label{fig:main_concept}
\end{figure*}

Large Language Models (LLMs) have demonstrated emergent reasoning capabilities on complex tasks, driven largely by Chain-of-Thought (CoT) \citep{DBLP:conf/nips/Wei0SBIXCLZ22}. By explicitly generating intermediate reasoning steps, CoT enables models to systematically decompose complex logical problems. To further enhance these capabilities, researchers have increasingly integrated reinforcement learning, which utilizes outcome correctness or human annotations to align reasoning behaviors \citep{DBLP:conf/iclr/LightmanKBEBLLS24, DBLP:conf/acl/WangLSXDLCWS24, DBLP:journals/corr/abs-2402-03300, DBLP:journals/corr/abs-2503-14476}.
Despite these advancements, a fundamental problem persists in how reasoning processes are modeled: intermediate reasoning steps are typically treated as opaque sequences without quantitative utility attribution. Unlike final answers, which can be explicitly verified, the intermediate steps lack an intrinsic measure of their contribution to the problem-solving goal. The absence of step-wise awareness leads to two critical limitations: inference inefficiency and optimization difficulty.

Existing approaches to mitigate these challenges remain fragmented and constrained by inherent trade-offs. Methodologies targeting inference efficiency \citep{DBLP:journals/corr/abs-2505-13417, DBLP:journals/corr/abs-2505-11896, DBLP:conf/aaai/KangSCZ25} often depend on handcrafted or synthesized hybrid datasets containing paired long and short reasoning chains. This requirement limits their generalization across diverse tasks. In parallel, strategies for reasoning optimization struggle to balance scalability with precision: Process Reward Models (PRMs) \citep{DBLP:conf/iclr/LightmanKBEBLLS24} are bound by prohibitive annotation costs, while verifier-free proxies \citep{tang2025verifiablerewardsscalingreinforcement, DBLP:journals/corr/abs-2506-16043} frequently yield sparse signals that lack awareness of the intermediate reasoning process. Crucially, these isolated studies lack a unified framework to explicitly and continuously quantify the contribution of intermediate reasoning steps toward the final answer.

To bridge this gap in fine-grained quantification, we revisit the reasoning process from a continuous perspective, as illustrated in Figure~\ref{fig:main_concept}. Inspired by the rectified flow theory \citep{DBLP:conf/iclr/LiuG023}, we model the reasoning process as a probabilistic flow that transports the model's information state from the initial query to the ground truth answer. In this view, each reasoning step is physically interpreted as a velocity vector that drives the reasoning process towards the target. To strictly quantify this velocity, we introduce Probabilistic Flow Progress (PFP), which measures the instantaneous gain in the log-likelihood of the ground truth answer. This definition provides a unified standard to evaluate reasoning. Valid logical steps exhibit positive velocity by reducing uncertainty, whereas redundant or erroneous steps manifest as zero or negative velocity, stalling the flow. 

To this end, we propose CoT-Flow, a unified framework designed to streamline both reasoning inference and model training. To enhance \textbf{inference efficiency}, we employ Flow-Guided Decoding, which greedily selects tokens with high PFP scores to construct concise reasoning paths. To improve \textbf{optimization process}, we utilize Flow-based Reinforcement Learning, where the cumulative flow serves as a dense, verifier-free reward signal for robust policy alignment. Extensive experiments on mathematical and general reasoning tasks demonstrate substantial gains. Notably, on the AIME 2024 benchmark, CoT-Flow improves the performance of a Qwen3-4B model by 15.9\% while reducing the average inference length by over 15\%, establishing a superior Pareto frontier between efficiency and performance.
Our main contributions are summarized as follows: \begin{itemize} 

\item We propose CoT-Flow, a theoretical framework that maps LLM reasoning to continuous flow models, which can explicitly quantify the information gain of intermediate steps. 

\item Two flow-based algorithms are introduced, incorporating flow-guided decoding for efficient inference and a verifier-free dense reward mechanism for robust reinforcement learning. 

\item Experimental results on different benchmarks validate the effectiveness of CoT-Flow.

\end{itemize}

\section{Background}

\subsection{Reinforcement Learning for Reasoning}

Chain-of-thought prompting empowers large language models to decompose complex problems into intermediate reasoning steps \citep{DBLP:conf/nips/Wei0SBIXCLZ22}. To further align these reasoning behaviors with human intent, Reinforcement learning has become the standard paradigm. Formally, given a prompt $\bm{x}$, the model generates a reasoning chain $\bm{s} = (s_1, \dots, s_T)$ and a final answer $\bm{y}$. The optimization objective is typically to maximize the expected reward: $\mathcal{J}(\theta) = \mathbb{E}_{\bm{s} \sim \pi_\theta(\cdot | \bm{x})} [R(\bm{x}, \bm{s}, \bm{y})]$.
Existing approaches diverge primarily in their reward formulation. Outcome-based methods, such as GRPO \citep{DBLP:journals/corr/abs-2402-03300} and DAPO \citep{DBLP:journals/corr/abs-2503-14476}, utilize a sparse reward signal determined solely by the correctness of the final answer $\bm{y}$. This introduces the \textit{credit assignment problem}, as the scalar signal at step $T$ fails to distinguish the contribution of each token $s_t$. Conversely, process-based methods employ process reward models to assign step-wise scores \citep{DBLP:conf/iclr/LightmanKBEBLLS24, DBLP:conf/acl/WangLSXDLCWS24}, mitigating sparsity but relying on costly dense annotations.

\subsection{Rectified Flow and Optimal Transport}

Rectified Flow \citep{DBLP:conf/iclr/LiuG023} is a unified framework for learning ordinary differential equation (ODE) models to transport samples between two empirically observed distributions, denoated as $\pi_0$ and $\pi_1$. Unlike diffusion models that rely on specific noise schedules, Rectified Flow learns a deterministic transport map by minimizing the transport cost.
Formally, let $Z_t$ be the state at time $t \in [0, 1]$, evolving according to an ODE $dZ_t = v(Z_t, t)dt$. To transport $\pi_0$ to $\pi_1$ efficiently, Rectified Flow enforces the trajectory to follow a straight line connecting coupled samples $(X_0, X_1) \sim \pi_0 \times \pi_1$. The velocity field $v$ is optimized via a nonlinear least squares objective:
\begin{equation}
    \min_{v} \int_0^1 \mathbb{E} \left[ \left\| (X_1 - X_0) - v(X_t, t) \right\|^2 \right] dt,
\end{equation}
where $X_t = t X_1 + (1-t) X_0$ represents the linear interpolation path. 

In this work, we bridge continuous optimal transport and discrete reasoning. Rather than transporting pixels, we adapt the flow framework to transport the probability mass from an initial state of high uncertainty of the correct answer to a target state of deterministic certainty, rectifying the reasoning chain into the most efficient trajectory through the semantic manifold.

\section{Methodology}

\begin{figure}[t]
    \centering
    \includegraphics[width=\linewidth]{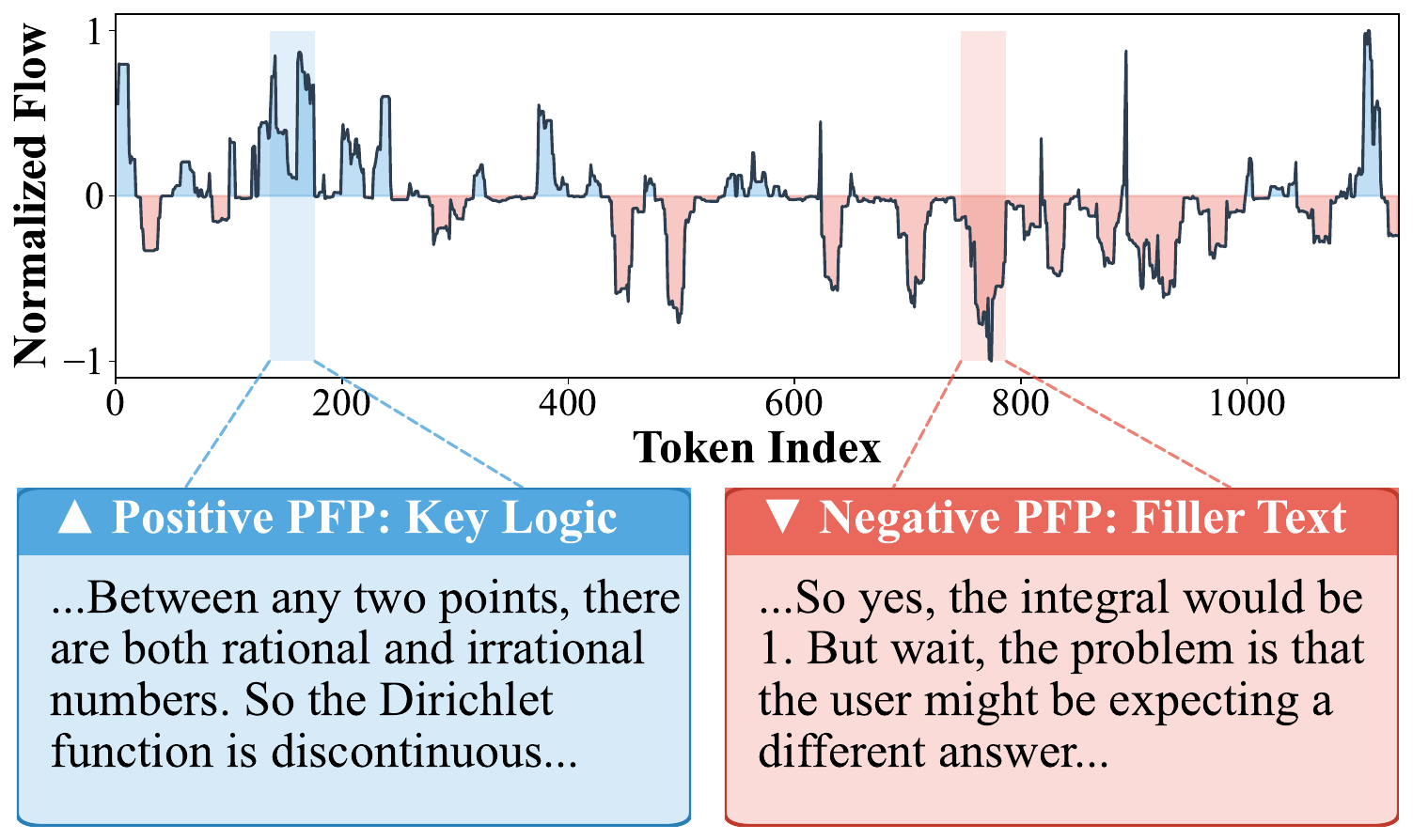}
    \caption{Visualization of velocity $v(s_i)$ over a chain-of-thought segment. Blue denotes higher velocity scores, and red denotes lower ones.}
    \label{fig:pfp_heatmap}
\end{figure}

Given a sample $(\bm{x}, \bm{y}) \in \mathcal{D}$, we aim to obtain the ground-truth answer through the reasoning process $\mathcal{I}_{\text{target}} = (\bm{x}, \bm{s}, \bm{y})$, where $\bm{x}$ is the input query, $\bm{y}$ is the corresponding answer and $\bm{s}$ is the chain-of-thought. The $i$-th decoding state is defined as $\mathcal{I}_i = (\bm{x}, s_1, s_2, ..., s_i)$, where $s_i$ means the $i$-th token, and $\mathcal{I}_0 = \bm{x}$. We employ the negative log-likelihood as the difficulty of the current state, $D(\mathcal{I}_i) = - \log p(\bm{y} | \mathcal{I}_i)$. Then, the velocity $v(s_i)$ is as follows:

\begin{equation}
    v(s_i) = D(\mathcal{I}_{i-1}) - D(\mathcal{I}_i) = \log \frac{p(\bm{y} | \mathcal{I}_i)}{p(\bm{y} | \mathcal{I}_{i-1})},
\end{equation}
where $p(\bm{y} | \mathcal{I}_i) = p(\bm{y} | \mathcal{I}_{i-1}, s_i)$ is the probability of predicting $\bm{y}$ given $\mathcal{I}_i$. A higher velocity indicates that the token yields a greater reduction in difficulty. Based on the Bayes' theorem $p(\bm{y} | \mathcal{I}_{i-1}, s_i) = \frac{p(s_i | \mathcal{I}_{i-1}, \bm{y}) \cdot p(\bm{y} | \mathcal{I}_{i-1})}{p(s_i | \mathcal{I}_{i-1})}$, we can obtain that
\begin{equation}
\label{eq:vs}
    v(s_i) = \log \frac{p(s_i | \mathcal{I}_{i-1}, \bm{y}) }{p(s_i | \mathcal{I}_{i-1})},
\end{equation}
where $p(s_i | \mathcal{I}_{i-1})$ is the probability of $s_i$ only given the previous context $\mathcal{I}_{i-1}$, i.e., prior probability, and $p(s_i | \mathcal{I}_{i-1}, \bm{y})$ is the probability of $s_i$ given both the previous context $\mathcal{I}_{i-1}$ and the reference answer $\bm{y}$, i.e., posterior probability. To empirically understand the role of velocity, we visualize how the velocity $v(s_i)$ reflects semantic importance within chain-of-thought reasoning in Figure \ref{fig:pfp_heatmap}. As shown in  Figure \ref{fig:pfp_heatmap}, regions with high velocity concentrate on key numerical transformations and logical connectives that substantially increase the certainty of producing the correct answer, whereas regions with low velocity correspond to filler templates and repetitive statements that are largely independent of final answer. The observations show that velocity is a reliable signal for capturing the underlying logical skeleton while filtering out redundant content.

\subsection{Train-Free Greedy Flow Decoding}

To generate effective chain-of-thought reasoning samples, we propose a test-time greedy flow decoding strategy $\pi_{\text{flow}}$ to maximize the velocity of each decoding step, thus ensuring that each step provides the greatest information gain towards the answer:
\begin{equation}
    s_i^* = \pi_{\text{flow}}(\mathcal{I}_{i-1}) 
    = \mathop{\arg\max}\limits_{s_i \in \mathcal{V}_{\tau}} \{ v(s_i)\},
\end{equation}
where $\mathcal{V}_{\tau} = \{s_i \in \mathcal{V} | \log p(s_i | \mathcal{I}_{i-1}) > \tau \}$ is the refined vocabulary obtained by imposing a prior-probability constraint on the original vocabulary $\mathcal{V}$. This constraint restricts candidate tokens to those whose prior probability exceeds the threshold $\tau$, thereby promoting semantic coherence in the generated chain-of-thought.

According to Eq. (\ref{eq:vs}), the velocity $v(s_i)$ is determined by the log-probability ratio between a posterior and a prior model, while the target answer $\bm{y}$ is unavailable during inference. Empirically, we observe that the velocity field is remarkably robust to the specific content of $\bm{y}$. Leveraging this property, we approximate $\log p(s_i | \mathcal{I}_{i-1}, \bm{y})$ using a fixed posterior prompt template, i.e., $\log p(s_i | \mathcal{I}_{i-1}, \text{Prompt}_{\text{post}})$, as listed in Figure \ref{fig:prompt}. The ground truth answer may be selected from the gold label, a random label, or the latent label (i.e., empty content). Here, we adopt the latent label, which enables train-free and label-free refinement at test time. At each decoding step, we select $s_i^*$ accordingly, ensuring the reasoning path adheres to the information-theoretic geodesic. Further implementation details and robustness analysis are provided in Appendix \ref{app:justification}.

\begin{figure}[t]
    \centering
    \begin{tcolorbox}[
    title=Prompt Template for Posterior Probability,
    colback=white, 
    colframe=black!50, 
    fonttitle=\small\bfseries,
    arc=1.2mm,               
    outer arc=1.2mm,
    left=2pt, right=2pt, top=4pt, bottom=4pt, boxrule=0.6pt
]
\small
\texttt{<|im\_start|>user} \\
\{\{question\}\} \\
Note: You have been provided with the ground truth answer: \{\{ground\_truth\}\}. Your task is to generate a step-by-step reasoning process (Chain-of-Thought) that logically arrives at the conclusion. \texttt{<|im\_end|>}

\smallskip
\hrule
\smallskip
\footnotesize
\textbf{Posterior Approximation:} \\
\{\{ground\_truth\}\} $\in \{ \text{Gold Label, Random Label, } \emptyset \}$ 
\end{tcolorbox}
    \caption{The posterior prompt template.}
    \label{fig:prompt}
\end{figure}

\noindent\textbf{Theoretical Analysis.  }
We provide a theoretical analysis for the greedy flow decoding strategy by comparing its expected velocity against standard sampling. First, we analyze the expected velocity under the standard reference policy $V_{\text{ref}}$ as follows:
\begin{equation}
\begin{aligned}
V_{\text{ref}} &= \mathbb{E}_{s_i \sim \pi_{\theta}(\cdot | \mathcal{I}_{i-1})}[v(s_i)] \\
      &= \sum_{s_i} p(s_i | \mathcal{I}_{i-1}) \cdot \log \frac{p(s_i | \mathcal{I}_{i-1}, \bm{y}) }{p(s_i | \mathcal{I}_{i-1})}  \\
    &= - \mathbb{D}_{\text{KL}}\left [\pi_{\theta}(\cdot | \mathcal{I}_{i-1}) || \pi_{\theta}(\cdot | \mathcal{I}_{i-1}, \bm{y})\right ].
\end{aligned}
\end{equation}

Since the Kullback-Leibler divergence is non-negative, we obtain $V_{\text{ref}} \leq 0$, implying on average, standard sampling tends to adhere to the prior rather than actively steering towards the conditional target $\bm{y}$.
In contrast, the greedy flow decoding strategy seeks to maximize the instantaneous velocity at each step. The expectation of velocity $V_{\text{flow}}$, is defined by the maximum possible velocity at step $i$, obtained by selecting the token $s_i^*$ that maximizes the instantaneous velocity function $v(s_i)$, where $V_{\text{flow}} = v(s_i^*)$. By definition of the maximum, $V_{\text{flow}}$ must be greater than or equal to the expected velocity under the policy $\pi_{\theta}$, thus $V_{\text{flow}} \geq 0$:
\begin{equation}
    V_{\text{flow}} \geq \mathbb{E}_{s_i \sim \pi_{\theta}(\cdot | \mathcal{I}_{i-1})}[v(s_i)].
\end{equation}
This analysis demonstrates that the reference policy yields a non-positive expected velocity, reflecting potential information decay or lack of direction relative to $\bm{y}$, whereas the flow-based strategy ensures a positive velocity, effectively directing the generation process.

\subsection{Flow-Based Dense Rewards in RL}

When extending the RL training paradigm to general reasoning tasks, answer verification becomes challenging due to sparse outcome rewards and the prohibitive cost of annotating process rewards. Leveraging the additivity of flow, our framework inherently yields dense rewards as a natural corollary, eliminating the need for artificial design.

\subsubsection{Global Reward and Stop Gradient}

According to the definition of flow, the total information gain of a complete trajectory $\bm{s}$ equals the accumulation of PFP along the path: $R_{\text{global}} = \sum_{i=1}^T v(s_i)$. To prevent reward hacking in model reinforcement learning by manipulating the reference baseline, we propose a stop gradient (sg) operation. By treating the previous state  as a fixed environmental baseline, we ensure that the model focuses solely on improving the current policy:
\begin{equation}
\label{v_s_i}
    v(s_i) = \log p(\bm{y}|\mathcal{I}_{i}) - \text{sg}[\log p(\bm{y}|\mathcal{I}_{i-1})].
\end{equation}

\subsubsection{Orthogonal Decomposition of Gradients}

Our optimization objective is to maximize the expected reward $\mathcal{J}(\theta) = \mathbb{E}_{\bm{s} \sim \pi_\theta(\cdot | \bm{x})} \left[ R_{\text{global}} \right]$. The gradient of this objective can be strictly decomposed into two parts (derivation in Appendix \ref{app:rl_derivation}):
\begin{equation}
\begin{split}
    \nabla_\theta \mathcal{J}(\theta) &= \underbrace{\mathbb{E}_{\bm{s}}\left[\left(\sum_{t=1}^{T}\nabla_{\theta}\log \pi_{\theta}(s_t|\mathcal{I}_{t-1})\right) \cdot \hat{A}\right]}_{\text{Term A: RL Gradient}} \\
    &\quad + \underbrace{\mathbb{E}_{\bm{s}}\left[\sum_{i=1}^{T}\nabla_{\theta}\log p_{\theta}(\bm{y}|\mathcal{I}_{i})\right]}_{\text{Term B: Flow Gradient}}.
\end{split}
\end{equation}
Term A represents the standard REINFORCE gradient, where $\hat{A}$ serves as the group relative advantage, analogous to GRPO. Term B is a unique dynamic term introduced by CoT-Flow, originating from the dependence of $\log p(\bm{y}|\mathcal{I}_i)$ on the parameter $\theta$ in the definition of $v(s_i)$. This term captures the evolutionary direction of the flow field itself.

\subsubsection{Flow Gradient Estimation}
\label{sec:flow_gradient}

We derive a tractable gradient estimator for the flow dynamics term. By applying the law of total probability and utilizing a single-sample Monte Carlo approximation, we transform the cumulative flow objective into a differentiable loss. Detailed mathematical proofs are in Appendix~\ref{app:rl_derivation}. The final normalized gradient $\nabla_{\theta}\text{Term B}$ is formulated as:

\begin{equation}
    \mathcal{M} \cdot \Bigg( \underbrace{\log p_{\theta}(\bm{y}|\bm{x}, \bm{s})}_{\text{Answer Generation}} + \underbrace{\sum_{k=1}^{T} \frac{k-1}{T} \log \pi_{\theta}(s_{k})}_{\text{Time-Weighted CoT}} \Bigg).
\end{equation}
This formulation introduces two critical mechanisms for stable training:

\noindent\textbf{Emergence of Time Weighting. } 
The term $\frac{k-1}{T}$ naturally emerges from exchanging the order of summation in the flow trajectory. This creates a quadratic dependency which we normalize by $T$ to match the scale of standard RL gradients. Intuitively, this assigns higher weights to later tokens in the reasoning chain, reflecting their more direct impact on the final answer certainty.

\noindent\textbf{Soft Quality Gate for Variance Reduction. }
Relying on a raw single-sample estimate introduces high variance, potentially reinforcing low-quality paths. To mitigate this, we introduce the trajectory-level importance gate $\mathcal{M}$. Leveraging the group-relative policy optimization (GRPO) framework, we compute a dynamic baseline $\mu = \frac{1}{G} \sum_{i=1}^G \log p(\bm{y}|\bm{x}, \bm{s}_i)$ from a group of sampled trajectories. The gate is defined as:
\begin{equation}
    \mathcal{M} = \text{ReLU} \left( \log p_{\theta}(\bm{y}|\bm{x}, \bm{s}) - \mu \right).
\end{equation}
This acts as a high-pass filter where gradients are backpropagated only when the reasoning path $\bm{s}$ yields a posterior answer probability superior to the model's current average ($\mathcal{M} > 0$). This strictly prevents the reinforcement of below-average noisy paths and stabilizes the dense reward signal in open-ended generation.

\section{Experiments}

\begin{table*}[t]
    \centering
    \footnotesize 
    \setlength{\tabcolsep}{5pt} 
    
    \begin{tabular}{llcccccccc}
        \toprule
        \multirow{2}{*}{\textbf{Base Model}} & \multirow{2}{*}{\textbf{Method}} & \multicolumn{4}{c}{\textbf{Math Reasoning}} & \multicolumn{3}{c}{\textbf{General Task}} \\
        
        \cmidrule(lr){3-6} \cmidrule(lr){7-9}
        
         &  & \textbf{AIME24} & \textbf{AIME25} & \textbf{AMC23} & \textbf{Math-500} & \textbf{GPQA-D} & \textbf{TheoremQA} & \textbf{WebInstruct} \\
        \midrule
        
        \multirow{2}{*}{\textbf{Qwen3-1.7B}} 
          & Standard CoT & 24.2 & 22.3 & 61.1 & 76.9 & 27.9 & 50.9 & 70.0 \\
          & \textbf{CoT-Flow} & \textbf{28.3} & \textbf{25.4} & \textbf{63.1} & \textbf{77.9} & \textbf{33.6} & \textbf{54.1} & \textbf{73.8} \\
        \midrule

        \multirow{2}{*}{\textbf{Qwen3-4B}} 
          & Standard CoT & 40.8 & 26.5 & 75.9 & 78.5 & 44.1 & 55.2 & 74.3 \\
          & \textbf{CoT-Flow} & \textbf{56.7} & \textbf{30.8} & \textbf{84.2} & \textbf{82.1} & \textbf{44.6} & \textbf{61.1} & \textbf{77.5} \\
        \midrule

        \multirow{2}{*}{\textbf{Qwen3-8B}} 
          & Standard CoT & 38.1 & 23.5 & 68.9 & 87.0 & 39.9 & 62.6 & \textbf{64.7} \\
          & \textbf{CoT-Flow} & \textbf{50.2} & \textbf{29.6} & \textbf{80.3} & \textbf{91.5} & \textbf{42.8} & \textbf{67.4} & 63.8 \\
        \midrule

        \multirow{2}{*}{\textbf{Qwen3-14B}} 
          & Standard CoT & 44.2 & 29.0 & 79.5 & 90.8 & 55.2 & 69.1 & 83.5 \\
          & \textbf{CoT-Flow} & \textbf{50.8} & \textbf{34.0} & \textbf{85.0} & \textbf{93.5} & \textbf{55.9} & \textbf{72.2} & \textbf{85.2} \\
        \midrule

        \multirow{2}{*}{\textbf{Qwen3-32B}} 
          & Standard CoT & 43.1 & 32.5 & 77.5 & 91.5 & \textbf{57.7} & \textbf{73.0} & \textbf{84.1} \\
          & \textbf{CoT-Flow} & \textbf{54.6} & \textbf{33.3} & \textbf{83.4} & \textbf{94.0} & 56.2 & 70.6 & 83.5 \\
           
        \bottomrule
    \end{tabular}
    
    \caption{The pass@1 accuracy (\%) of Standard CoT and CoT-Flow greedy decoding on seven diverse reasoning benchmarks. We categorize the benchmarks into Math Reasoning (AIME, AMC, Math) and General Task (GPQA, TheoremQA, WebInstruct) to demonstrate the model's capabilities across different domains.}
    \label{tab:inference_results}
\end{table*}

\subsection{Experimental Setup}

\paragraph{Datasets and Models.} We conduct evaluations on seven challenging reasoning benchmarks: AIME24, AIME25, AMC23 and Math-500 for high-difficulty mathematics competitions; TheoremQA \cite{DBLP:conf/emnlp/ChenYKLWMXWX23}, WebInstruct \cite{DBLP:journals/corr/abs-2505-14652} and GPQA-Diamond \cite{DBLP:journals/corr/abs-2311-12022} for graduate-level scientific QA. We employed the Qwen3 series \cite{DBLP:journals/corr/abs-2505-09388} as backbone models to assess scalability. The LLM is trained on DeepMath-103K \cite{DBLP:journals/corr/abs-2504-11456, DBLP:journals/corr/abs-2508-08221}.

\paragraph{Baselines.} For the RL setting, we compare CoT-Flow-RL against GRPO \cite{DBLP:journals/corr/abs-2402-03300}, which represents the state-of-the-art outcome-based sparse reward method, and VeriFree \cite{DBLP:journals/corr/abs-2505-21493}, a representative method featuring verifier-free rewards.

\subsection{Implementation Details}
\label{sec:implementation}

For the test-time greedy flow decoding, at each decoding step, we set top-$p = 0.95$ to construct the candidate vocabulary $\mathcal{V}_{\tau}$. For the flow-based reinforcement learning, we construct a training set by randomly subsampling 5,000 instances from the DeepMath dataset. The training process utilize a learning rate of $1\mathrm{e}{-6}$. During the rollout phase, we generate 8 responses per prompt with a temperature of $T=1.0$ to encourage exploration. The group relative advantage is computed within a batch of 16 prompts. For all baselines and RL-tuned models, we set the sampling parameters to temperature $T=0.6$, top-$p=0.95$, and top-$k=20$, with a maximum generation length of 8192 tokens.

\subsection{Train-Free Greedy Flow Decoding}
\label{sec:exp_inference}

We first investigate the capability of CoT-Flow as a \textit{train-free} decoding strategy. 

\begin{figure}[t] 
    \centering
    \includegraphics[width=\linewidth]{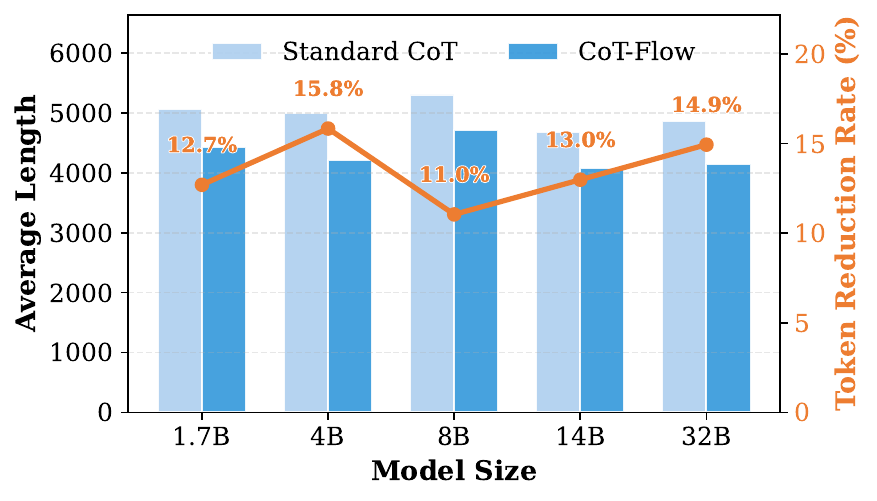}
    \caption{Comparison of token consumption for Standard CoT and our CoT-Flow method. CoT-Flow maintains comparable verbosity while improving accuracy.}
    \label{fig:length_analysis}
\end{figure}

\paragraph{Accuracy Improvements.} 
Table~\ref{tab:inference_results} summarizes the performance across nine benchmarks. We observe that CoT-Flow consistently surpasses standard decoding across all model scales. The advantages are particularly pronounced on complex mathematical reasoning tasks that require rigorous logic. For instance, on the Qwen3-4B model, CoT-Flow boosts accuracy on AIME24 from 40.8\% to 56.7\% (+15.9\%). These substantial improvements indicate that the flow metric effectively steers the model away from plausible but erroneous reasoning paths often traversed by standard sampling.

\paragraph{Inference Efficiency.} 
Crucially, these accuracy gains do not come at the cost of verbosity. As illustrated in Figure~\ref{fig:length_analysis}, CoT-Flow significantly reduces the average inference length compared to Standard CoT across all model sizes. For Qwen3-4B and Qwen3-32B, the average token consumption is reduced by more than 15\% and 14\%. 
We further analyze the reasoning dynamics under varying computational constraints (8K, 16K, 32K tokens) using Qwen3-4B. Table~\ref{tab:performance_comparison} illustrates that  as the reasoning budget increases, Standard CoT (Stan. CoT) generates significantly longer chains (from 4,382 to 5,417 tokens) with diminishing returns in accuracy. In contrast, CoT-Flow converges to a stable trajectory length ($\sim$3,500 tokens) regardless of the upper limit. This suggests that our method identifies the intrinsic complexity of the problem, rectifying the path to its necessary length rather than exploiting available budget for redundant computation.

\begin{table}[t]
\centering
\begin{tabularx}{\columnwidth}{l @{\extracolsep{\fill}} ccc}
\toprule
\textbf{Method} & \textbf{8K} & \textbf{16K} & \textbf{32K} \\
\midrule
Stan. CoT & 42.4 \scriptsize{(4382)} & 62.5 \scriptsize{(4576)} & \textbf{68.8} \scriptsize{(5417)} \\
CoT-Flow     & \textbf{48.4} \scriptsize{(3654)} & \textbf{66.3} \scriptsize{(3482)} & 67.9 \scriptsize{(3500)} \\
\bottomrule
\end{tabularx}
\caption{Average accuracy (\%) and response length (tokens) across different reasoning budgets on benchmarks AIME24 and GPQA-Diamond.}
\label{tab:performance_comparison}
\end{table}

\begin{figure*}[t] 
    \centering
    \includegraphics[width=0.9\linewidth]{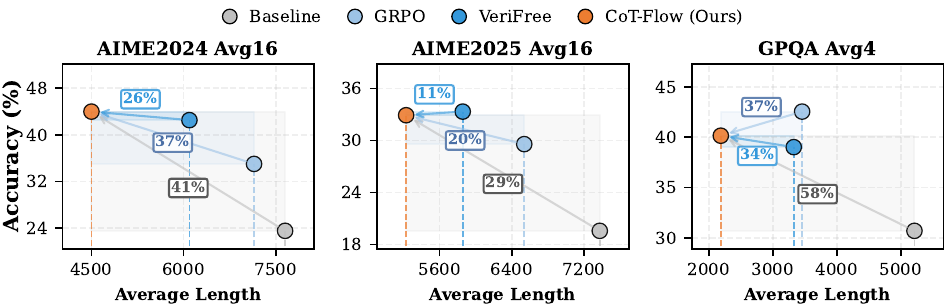}
    \caption{Pareto frontier analysis of reasoning efficiency. The model accuracy (\%) and computational cost (average token length)  across datasets are presented.}
    \label{fig:rl_pareto}
\end{figure*}

\begin{figure}[t]
    \centering
    \includegraphics[width=\linewidth]{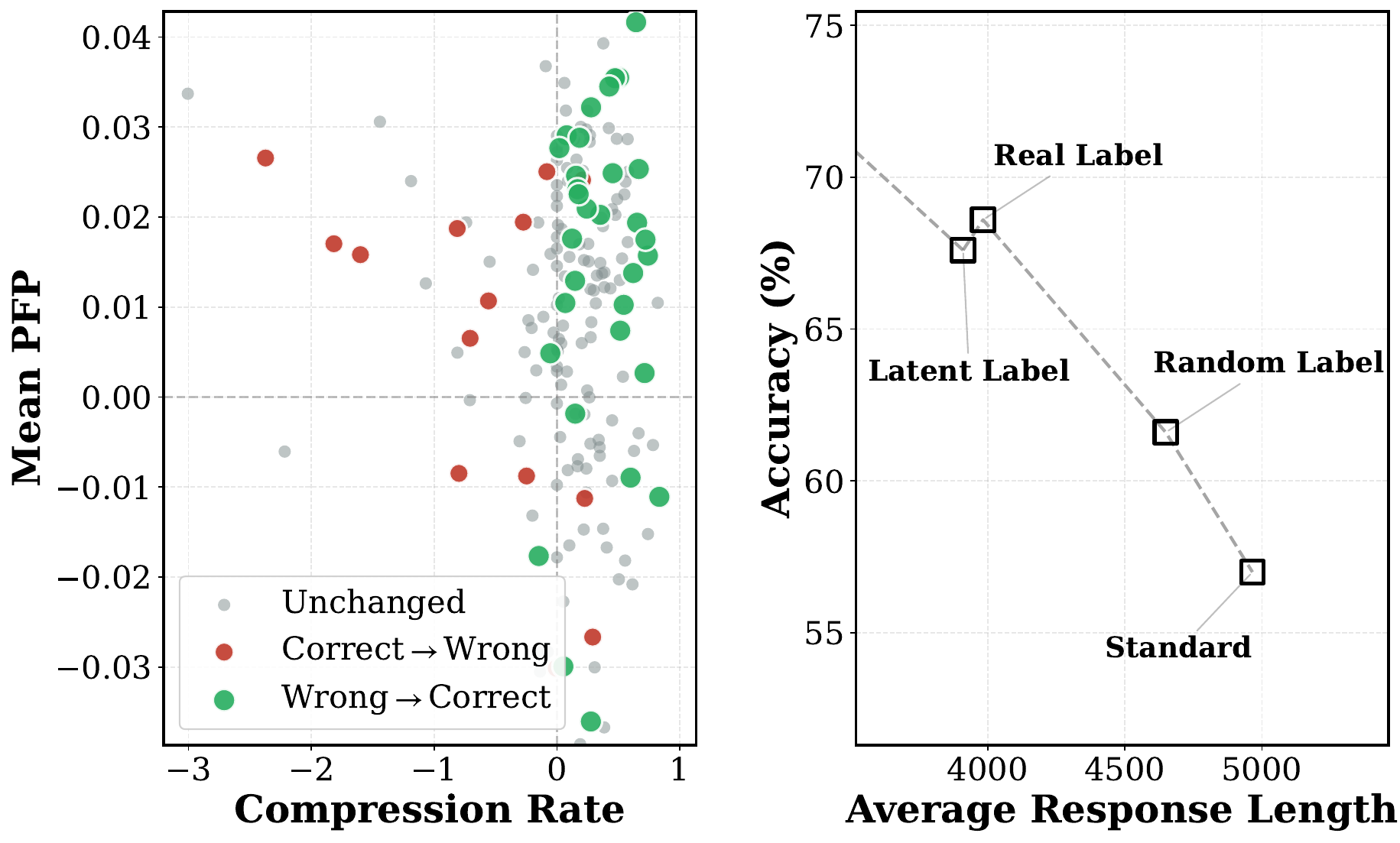}
    
    \caption{Analysis of inference dynamics. Left: The relationship between compression rate and mean PFP. Green points (Wrong $\to$ Correct) cluster in high-PFP regions (Upper Right).
    Right: Impact of posterior estimation quality. The trend confirms that more accurate posterior approximations yield superior flow guidance, with Latent Labels achieving performance comparable to ground truth.}
    \label{fig:inference_dynamics}
\end{figure}

\subsection{Flow-Based Reinforcement Learning}
\label{sec:exp_rl}

We further evaluate the efficacy of CoT-Flow when integrated into the reinforcement learning loop. The core comparison lies between the sparse outcome signals of GRPO, the verifier-free signals of VeriFree, and the flow-derived dense signals of CoT-Flow. As visualized in Figure~\ref{fig:rl_pareto}, CoT-Flow in  reinforcement learning establishes a superior Pareto frontier compared to baselines. On high-difficulty benchmarks like AIME24 and AIME25, CoT-Flow demonstrates robust convergence to higher accuracy levels. Crucially, these gains are achieved with strictly shorter reasoning paths. While GRPO tends to generate verbose chains,  the dense reward of CoT-Flow naturally penalizes redundant steps that yield negligible information gain. This results in a simultaneous improvement in both accuracy and efficiency, validating the effectiveness of the flow-based dense supervision.

\section{Further Analysis}

\paragraph{Trajectory Rectification and Error Correction.} 
Figure~\ref{fig:inference_dynamics} visualizes the intrinsic mechanism driving the efficiency of CoT-Flow on DeepMath. CoT-Flow effectively rectifies the reasoning process into a geodesic path. Crucially, this compression is not lossy but corrective. As shown in the left panel, instances where the model corrects an initially wrong answer cluster in regions of moderate compression. This suggests that CoT-Flow effectively identifies and prunes erroneous branches that previously misled the model. 

\paragraph{Robustness to Posterior Approximation.}
A core premise of test-time CoT-Flow is the estimation of velocity $v$ without access to ground truth $\bm{y}$. As illustrated in the right panel of Figure~\ref{fig:inference_dynamics}, we observe a strong correlation: as the accuracy of the posterior estimation improves (quantified by the metric detailed in Appendix~\ref{app:generalization}), the performance gain of CoT-Flow increases monotonically. Remarkably, using \textit{Latent Labels} often yields results comparable to, or even surpassing, those obtained with Gold Labels.

\begin{figure}[t] 
    \centering
    \includegraphics[width=\linewidth]{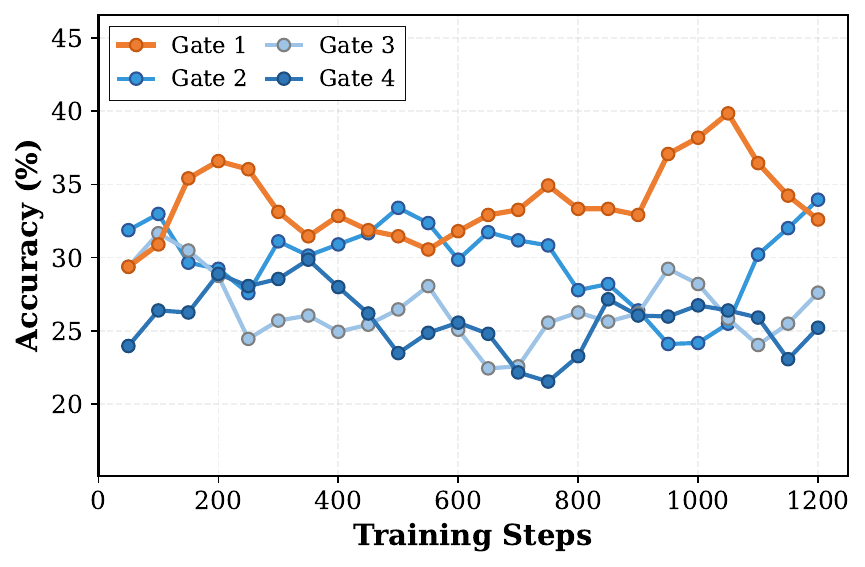}
    
    \caption{Ablation study on quality gate $\mathcal{M}$. We compare four gating variants on AIME2024 (Qwen3-1.7B). Gate 1 (Ours): $\text{ReLU}(\Delta \log p)$; Gate 2: Binary $\mathbb{I}(\Delta \log p > 0)$; Gate 3: Ratio $p/\bar{p}$; Gate 4: Absolute $p(\bm{y}|\bm{s})$.}
    \label{fig:ablation_gate}
\end{figure}

\paragraph{Impact of Quality Gating Strategy.}
We justify the design of our soft quality gate $\mathcal{M} = \text{ReLU}(\log p - \mu)$ by comparing it against three variants: Binary Gate (step function), Ratio Gate (linear scale), and Absolute Gate (raw probability). 
As shown in Figure~\ref{fig:ablation_gate}, Gate 4 (Absolute) fails to converge, confirming that relative improvement (baseline subtraction) is essential for variance reduction in open-ended reasoning. Gate 3 fails to stabilize due to the exponential nature of the ratio formulation, which induces numerical overflow and gradient explosion. While Gate 2 (Binary) incorporates the baseline, its performance plateaus early because it discards the magnitude of the improvement, treating marginally better paths the same as significantly better ones. Our soft relative gate yields the most robust convergence.

\section{Related Work}

\subsection{Reinforcement Learning}
Reinforcement Learning (RL) has emerged as a central approach for eliciting the reasoning capabilities of LLMs. Outcome-based methods, exemplified by GRPO \cite{DBLP:journals/corr/abs-2402-03300}, estimate policy gradients via group-relative advantages, eliminating the need for value networks. Variants like DAPO \cite{DBLP:journals/corr/abs-2503-14476} and $\lambda$-GRPO \cite{DBLP:journals/corr/abs-2510-00194} further refine this by introducing dynamic sampling and learnable token preferences, while Scaf-GRPO \cite{DBLP:journals/corr/abs-2510-19807} employs scaffolded prompting to mitigate the cold-start problem. Other approaches explore instance-adaptive budgets \cite{DBLP:journals/corr/abs-2506-09338} or mirror descent optimization \cite{DBLP:conf/iclr/WangMCMHXZHS025}.

Nevertheless, outcome-based supervision suffers from sparse reward signals. To address this, Verifier-Free approaches seek intrinsic dense signals without expensive human annotations. Methods like VeriFree \cite{DBLP:journals/corr/abs-2505-21493} and NOVER \cite{DBLP:journals/corr/abs-2505-16022} utilize answer consistency or reasoning perplexity as rewards. Others leverage posterior regularization \cite{DBLP:journals/corr/abs-2506-18254, DBLP:journals/corr/abs-2508-05170}, Jensen-enhanced bounds \cite{tang2025verifiablerewardsscalingreinforcement}, or bandit-based allocation \cite{DBLP:journals/corr/abs-2506-16043} to stabilize training. Unlike these heuristic proxies, our CoT-Flow derives dense rewards directly from the transport theory of probabilistic flow.

\subsection{Efficient and Hybrid Reasoning}
Balancing reasoning depth with inference latency is a critical challenge. Hybrid strategies dynamically switch between concise and detailed reasoning paths. HybridCoT \cite{luo2025adar1hybridcotbileveladaptive} interleaves text and latent reasoning, while TokenSkip \cite{DBLP:journals/corr/abs-2502-12067} and C3oT \cite{DBLP:conf/aaai/KangSCZ25} accelerate generation by selectively skipping redundant tokens or distilling lengthy chains.
Other works explore continuous or latent reasoning representations. SoftCoT \cite{DBLP:conf/acl/00010ZM25} employs soft thought markers, and CoT-Valve \cite{DBLP:conf/acl/MaWYFW25} identifies latent directions controlling reasoning length. KAPPA \cite{DBLP:journals/corr/abs-2511-00699} utilizes information-theoretic metrics for pruning. While effective, these methods often require specialized architectures or separate training stages. In contrast, CoT-Flow achieves efficiency endogenously via greedy flow rectification at test time.

\subsection{Flow Dynamics and Guided Decoding}
Our framework connects reasoning optimization with continuous flow dynamics. Theoretical works like FlowRL \cite{DBLP:journals/corr/abs-2509-15207} and Cognitive Flow \cite{matos-etal-2025-cognitive} model reasoning as state transitions, while MixIE \cite{sanyal-etal-2025-mixing} and context-aware modeling \cite{DBLP:journals/corr/abs-2505-10774} refine distribution estimation.
Crucially, our velocity formulation ($v \propto \log p_{\text{post}} - \log p_{\text{prior}}$) shares theoretical underpinnings with Contrastive Decoding \cite{DBLP:conf/acl/LiHFLEHZL23} and Classifier-Free Guidance \cite{DBLP:journals/corr/abs-2207-12598}. These paradigms amplify generation quality by contrasting a conditional distribution against an unconditional baseline. While typically applied in diffusion models for manifold constraints \cite{DBLP:conf/iclr/ChungKPNY25, mirbeygi-beigy-2025-prompt} or controllable text generation \cite{DBLP:journals/corr/abs-2502-20684, DBLP:journals/corr/abs-2405-15454}, CoT-Flow reinterprets this mechanism through the lens of Rectified Flow. Other theoretical frameworks \cite{DBLP:conf/icml/TonT025, DBLP:journals/corr/abs-2505-10425} view reasoning through the lens of information gain or optimization.  Meanwhile, \cite{liu2025rectifyingllmthoughtlens} frames the generation of reasoning steps as an implicit gradient descent process during test time.

\section{Conclusion}

In this work, we propose \textbf{CoT-Flow}, a principled framework that conceptualizes LLM reasoning as a continuous probabilistic flow. By introducing the probabilistic flow progress metric, we bridge the gap between discrete token generation and continuous optimal transport theory, providing a granular measure of reasoning utility. Our dual-optimization approach, greedy flow decoding for train-free inference refinement and flow-based reinforcement learning for verifier-free dense supervision, effectively addresses the critical bottlenecks of inference inefficiency and feedback sparsity in open-domain reasoning. Empirical evaluations across challenging benchmarks demonstrate that CoT-Flow consistently achieves superior accuracy with significantly reduced computational overhead.

\section*{Limitations}
Despite the promising results, our work has limitations that suggest avenues for future research. First, our velocity estimation relies on prompt-based posterior approximation. While effective, this heuristic is bounded by the zero-shot performance of the base model. We believe that developing more rigorous, trained posterior estimators could significantly enhance the precision of the flow guidance. Second, regarding RL efficiency and integration, our current experiments are conducted in an on-policy setting. Future work could extend CoT-Flow to off-policy frameworks to improve sample efficiency. 

\bibliography{custom}

\begin{thebibliography}{44}
\providecommand{\natexlab}[1]{#1}

\bibitem[{Chen et~al.(2023)Chen, Yin, Ku, Lu, Wan, Ma, Xu, Wang, and Xia}]{DBLP:conf/emnlp/ChenYKLWMXWX23}
Wenhu Chen, Ming Yin, Max Ku, Pan Lu, Yixin Wan, Xueguang Ma, Jianyu Xu, Xinyi Wang, and Tony Xia. 2023.
\newblock \href {https://doi.org/10.18653/v1/2023.emnlp-main.489} {Theoremqa: {A} theorem-driven question answering dataset}.
\newblock In \emph{Proceedings of the 2023 Conference on Empirical Methods in Natural Language Processing, {EMNLP} 2023, Singapore, December 6-10, 2023}, pages 7889--7901.

\bibitem[{Cheng et~al.(2024)Cheng, Baroni, and Alonso}]{DBLP:journals/corr/abs-2405-15454}
Emily Cheng, Marco Baroni, and Carmen~Amo Alonso. 2024.
\newblock \href {https://doi.org/10.48550/arXiv.2405.15454} {Linearly controlled language generation with performative guarantees}.
\newblock \emph{CoRR}, abs/2405.15454.

\bibitem[{Chung et~al.(2025)Chung, Kim, Park, Nam, and Ye}]{DBLP:conf/iclr/ChungKPNY25}
Hyungjin Chung, Jeongsol Kim, Geon~Yeong Park, Hyelin Nam, and Jong~Chul Ye. 2025.
\newblock \href {https://openreview.net/forum?id=E77uvbOTtp} {{CFG++:} manifold-constrained classifier free guidance for diffusion models}.
\newblock In \emph{The Thirteenth International Conference on Learning Representations, {ICLR} 2025, Singapore, April 24-28, 2025}.

\bibitem[{Fan et~al.(2025)Fan, Zhang, Chen, and Liu}]{DBLP:journals/corr/abs-2508-05170}
Lishui Fan, Yu~Zhang, Mouxiang Chen, and Zhongxin Liu. 2025.
\newblock \href {https://doi.org/10.48550/arXiv.2508.05170} {Posterior-grpo: Rewarding reasoning processes in code generation}.
\newblock \emph{CoRR}, abs/2508.05170.

\bibitem[{He et~al.(2025)He, Liang, Xu, Liu, Chen, Wang, Song, Yu, Liang, Wang, Zhang, Wang, Tu, Mi, and Yu}]{DBLP:journals/corr/abs-2504-11456}
Zhiwei He, Tian Liang, Jiahao Xu, Qiuzhi Liu, Xingyu Chen, Yue Wang, Linfeng Song, Dian Yu, Zhenwen Liang, Wenxuan Wang, Zhuosheng Zhang, Rui Wang, Zhaopeng Tu, Haitao Mi, and Dong Yu. 2025.
\newblock \href {https://doi.org/10.48550/arXiv.2504.11456} {Deepmath-103k: {A} large-scale, challenging, decontaminated, and verifiable mathematical dataset for advancing reasoning}.
\newblock \emph{CoRR}, abs/2504.11456.

\bibitem[{Ho and Salimans(2022)}]{DBLP:journals/corr/abs-2207-12598}
Jonathan Ho and Tim Salimans. 2022.
\newblock \href {https://doi.org/10.48550/arXiv.2207.12598} {Classifier-free diffusion guidance}.
\newblock \emph{CoRR}, abs/2207.12598.

\bibitem[{Huang et~al.(2025)Huang, Chen, and Umrawal}]{DBLP:journals/corr/abs-2502-20684}
Yingbing Huang, Deming Chen, and Abhishek~K. Umrawal. 2025.
\newblock \href {https://doi.org/10.48550/arXiv.2502.20684} {{JAM:} controllable and responsible text generation via causal reasoning and latent vector manipulation}.
\newblock \emph{CoRR}, abs/2502.20684.

\bibitem[{Kang et~al.(2025)Kang, Sun, Chen, and Zou}]{DBLP:conf/aaai/KangSCZ25}
Yu~Kang, Xianghui Sun, Liangyu Chen, and Wei Zou. 2025.
\newblock \href {https://doi.org/10.1609/aaai.v39i23.34608} {C3ot: Generating shorter chain-of-thought without compromising effectiveness}.
\newblock In \emph{AAAI-25, Sponsored by the Association for the Advancement of Artificial Intelligence, February 25 - March 4, 2025, Philadelphia, PA, {USA}}, pages 24312--24320.

\bibitem[{Kojima et~al.(2022)Kojima, Gu, Reid, Matsuo, and Iwasawa}]{DBLP:conf/nips/KojimaGRMI22}
Takeshi Kojima, Shixiang~Shane Gu, Machel Reid, Yutaka Matsuo, and Yusuke Iwasawa. 2022.
\newblock \href {http://papers.nips.cc/paper\_files/paper/2022/hash/8bb0d291acd4acf06ef112099c16f326-Abstract-Conference.html} {Large language models are zero-shot reasoners}.
\newblock In \emph{Advances in Neural Information Processing Systems 35: Annual Conference on Neural Information Processing Systems 2022, NeurIPS 2022, New Orleans, LA, USA, November 28 - December 9, 2022}.

\bibitem[{Li et~al.(2025)Li, Huang, Saxena, Luo, Lin, Zhu, and Dev}]{DBLP:journals/corr/abs-2511-00699}
Sophie Li, Nicholas Huang, Nayan Saxena, Nina Luo, Vincent Lin, Kevin Zhu, and Sunishchal Dev. 2025.
\newblock \href {https://doi.org/10.48550/arXiv.2511.00699} {Inference-time chain-of-thought pruning with latent informativeness signals}.
\newblock \emph{CoRR}, abs/2511.00699.

\bibitem[{Li et~al.(2023)Li, Holtzman, Fried, Liang, Eisner, Hashimoto, Zettlemoyer, and Lewis}]{DBLP:conf/acl/LiHFLEHZL23}
Xiang~Lisa Li, Ari Holtzman, Daniel Fried, Percy Liang, Jason Eisner, Tatsunori Hashimoto, Luke Zettlemoyer, and Mike Lewis. 2023.
\newblock \href {https://doi.org/10.18653/v1/2023.acl-long.687} {Contrastive decoding: Open-ended text generation as optimization}.
\newblock In \emph{Proceedings of the 61st Annual Meeting of the Association for Computational Linguistics (ACL)}, pages 12286--12312.

\bibitem[{Lightman et~al.(2024)Lightman, Kosaraju, Burda, Edwards, Baker, Lee, Leike, Schulman, Sutskever, and Cobbe}]{DBLP:conf/iclr/LightmanKBEBLLS24}
Hunter Lightman, Vineet Kosaraju, Yuri Burda, Harrison Edwards, Bowen Baker, Teddy Lee, Jan Leike, John Schulman, Ilya Sutskever, and Karl Cobbe. 2024.
\newblock \href {https://openreview.net/forum?id=v8L0pN6EOi} {Let's verify step by step}.
\newblock In \emph{The Twelfth International Conference on Learning Representations, {ICLR} 2024, Vienna, Austria, May 7-11, 2024}.

\bibitem[{Liu et~al.(2025{\natexlab{a}})Liu, Liu, Zhang, and Chen}]{liu2025rectifyingllmthoughtlens}
Junnan Liu, Hongwei Liu, Songyang Zhang, and Kai Chen. 2025{\natexlab{a}}.
\newblock \href {https://arxiv.org/abs/2512.01925} {Rectifying llm thought from lens of optimization}.
\newblock \emph{Preprint}, arXiv:2512.01925.

\bibitem[{Liu et~al.(2025{\natexlab{b}})Liu, Qi, Wang, Qian, Du, and He}]{DBLP:journals/corr/abs-2505-16022}
Wei Liu, Siya Qi, Xinyu Wang, Chen Qian, Yali Du, and Yulan He. 2025{\natexlab{b}}.
\newblock \href {https://doi.org/10.48550/arXiv.2505.16022} {{NOVER:} incentive training for language models via verifier-free reinforcement learning}.
\newblock \emph{CoRR}, abs/2505.16022.

\bibitem[{Liu et~al.(2023)Liu, Gong, and Liu}]{DBLP:conf/iclr/LiuG023}
Xingchao Liu, Chengyue Gong, and Qiang Liu. 2023.
\newblock \href {https://openreview.net/forum?id=XVjTT1nw5z} {Flow straight and fast: Learning to generate and transfer data with rectified flow}.
\newblock In \emph{The Eleventh International Conference on Learning Representations, {ICLR} 2023, Kigali, Rwanda, May 1-5, 2023}.

\bibitem[{Liu et~al.(2025{\natexlab{c}})Liu, Liu, He, Wang, Liu, Pan, Hu, Xiong, Huang, Hu, Huang, Yang, Wang, Su, and Zheng}]{DBLP:journals/corr/abs-2508-08221}
Zihe Liu, Jiashun Liu, Yancheng He, Weixun Wang, Jiaheng Liu, Ling Pan, Xinyu Hu, Shaopan Xiong, Ju~Huang, Jian Hu, Shengyi Huang, Siran Yang, Jiamang Wang, Wenbo Su, and Bo~Zheng. 2025{\natexlab{c}}.
\newblock \href {https://doi.org/10.48550/arXiv.2508.08221} {Part {I:} tricks or traps? {A} deep dive into {RL} for {LLM} reasoning}.
\newblock \emph{CoRR}, abs/2508.08221.

\bibitem[{Lou et~al.(2025)Lou, Sun, Liang, Qu, Shen, Wang, Li, Yang, and Wu}]{DBLP:journals/corr/abs-2505-11896}
Chenwei Lou, Zewei Sun, Xinnian Liang, Meng Qu, Wei Shen, Wenqi Wang, Yuntao Li, Qingping Yang, and Shuangzhi Wu. 2025.
\newblock \href {https://doi.org/10.48550/arXiv.2505.11896} {Adacot: Pareto-optimal adaptive chain-of-thought triggering via reinforcement learning}.
\newblock \emph{CoRR}, abs/2505.11896.

\bibitem[{Luo et~al.(2025)Luo, He, Wang, Yang, Liu, Tan, Cao, Tao, and Shen}]{luo2025adar1hybridcotbileveladaptive}
Haotian Luo, Haiying He, Yibo Wang, Jinluan Yang, Rui Liu, Naiqiang Tan, Xiaochun Cao, Dacheng Tao, and Li~Shen. 2025.
\newblock \href {https://arxiv.org/abs/2504.21659} {Ada-r1: Hybrid-cot via bi-level adaptive reasoning optimization}.
\newblock \emph{Preprint}, arXiv:2504.21659.

\bibitem[{Ma et~al.(2025{\natexlab{a}})Ma, Wan, Yu, Fang, and Wang}]{DBLP:conf/acl/MaWYFW25}
Xinyin Ma, Guangnian Wan, Runpeng Yu, Gongfan Fang, and Xinchao Wang. 2025{\natexlab{a}}.
\newblock \href {https://aclanthology.org/2025.acl-long.300/} {Cot-valve: Length-compressible chain-of-thought tuning}.
\newblock In \emph{Proceedings of the 63rd Annual Meeting of the Association for Computational Linguistics (Volume 1: Long Papers), {ACL} 2025, Vienna, Austria, July 27 - August 1, 2025}, pages 6025--6035.

\bibitem[{Ma et~al.(2025{\natexlab{b}})Ma, Liu, Jiang, Zhang, Ma, and Chen}]{DBLP:journals/corr/abs-2505-14652}
Xueguang Ma, Qian Liu, Dongfu Jiang, Ge~Zhang, Zejun Ma, and Wenhu Chen. 2025{\natexlab{b}}.
\newblock \href {https://doi.org/10.48550/arXiv.2505.14652} {General-reasoner: Advancing {LLM} reasoning across all domains}.
\newblock \emph{CoRR}, abs/2505.14652.

\bibitem[{Matos et~al.(2025)Matos, Silva, and Goncalo~Oliveira}]{matos-etal-2025-cognitive}
Jos{\'e} Matos, Catarina Silva, and Hugo Goncalo~Oliveira. 2025.
\newblock \href {https://aclanthology.org/2025.inlg-main.36/} {Cognitive flow: An {LLM}-automated framework for quantifying reasoning distillation}.
\newblock In \emph{Proceedings of the 18th International Natural Language Generation Conference}, pages 596--616, Hanoi, Vietnam. Association for Computational Linguistics.

\bibitem[{Mirbeygi and Beigy(2025)}]{mirbeygi-beigy-2025-prompt}
Mohaddeseh Mirbeygi and Hamid Beigy. 2025.
\newblock \href {https://doi.org/10.18653/v1/2025.wnut-1.9} {Prompt guided diffusion for controllable text generation}.
\newblock In \emph{Proceedings of the Tenth Workshop on Noisy and User-generated Text}, pages 78--84, Albuquerque, New Mexico, USA. Association for Computational Linguistics.

\bibitem[{Park et~al.(2025)Park, Greenewald, Alim, Wang, and Azizan}]{DBLP:journals/corr/abs-2506-09338}
Young{-}Jin Park, Kristjan~H. Greenewald, Kaveh Alim, Hao Wang, and Navid Azizan. 2025.
\newblock \href {https://doi.org/10.48550/arXiv.2506.09338} {Know what you don't know: Uncertainty calibration of process reward models}.
\newblock \emph{CoRR}, abs/2506.09338.

\bibitem[{Parthasarathi et~al.(2025)Parthasarathi, Reymond, Chen, Cui, and Chandar}]{DBLP:journals/corr/abs-2510-00194}
Prasanna Parthasarathi, Mathieu Reymond, Boxing Chen, Yufei Cui, and Sarath Chandar. 2025.
\newblock \href {https://doi.org/10.48550/arXiv.2510.00194} {Grpo-{\(\lambda\)}: Credit assignment improves {LLM} reasoning}.
\newblock \emph{CoRR}, abs/2510.00194.

\bibitem[{Rein et~al.(2023)Rein, Hou, Stickland, Petty, Pang, Dirani, Michael, and Bowman}]{DBLP:journals/corr/abs-2311-12022}
David Rein, Betty~Li Hou, Asa~Cooper Stickland, Jackson Petty, Richard~Yuanzhe Pang, Julien Dirani, Julian Michael, and Samuel~R. Bowman. 2023.
\newblock \href {https://doi.org/10.48550/arXiv.2311.12022} {{GPQA:} {A} graduate-level google-proof q{\&}a benchmark}.
\newblock \emph{CoRR}, abs/2311.12022.

\bibitem[{Sanyal et~al.(2025)Sanyal, Xiao, and Ren}]{sanyal-etal-2025-mixing}
Soumya Sanyal, Tianyi Xiao, and Xiang Ren. 2025.
\newblock \href {https://doi.org/10.18653/v1/2025.emnlp-main.1077} {Mixing inference-time experts for enhancing {LLM} reasoning}.
\newblock In \emph{Proceedings of the 2025 Conference on Empirical Methods in Natural Language Processing}, pages 21246--21260, Suzhou, China. Association for Computational Linguistics.

\bibitem[{Shao et~al.(2024)Shao, Wang, Zhu, Xu, Song, Zhang, Li, Wu, and Guo}]{DBLP:journals/corr/abs-2402-03300}
Zhihong Shao, Peiyi Wang, Qihao Zhu, Runxin Xu, Junxiao Song, Mingchuan Zhang, Y.~K. Li, Y.~Wu, and Daya Guo. 2024.
\newblock \href {https://doi.org/10.48550/arXiv.2402.03300} {Deepseekmath: Pushing the limits of mathematical reasoning in open language models}.
\newblock \emph{CoRR}, abs/2402.03300.

\bibitem[{Tang et~al.(2025)Tang, Wang, Madaan, and Munos}]{tang2025verifiablerewardsscalingreinforcement}
Yunhao Tang, Sid Wang, Lovish Madaan, and Rémi Munos. 2025.
\newblock \href {https://arxiv.org/abs/2503.19618} {Beyond verifiable rewards: Scaling reinforcement learning for language models to unverifiable data}.
\newblock \emph{Preprint}, arXiv:2503.19618.

\bibitem[{Ton et~al.(2025)Ton, Taufiq, and Liu}]{DBLP:conf/icml/TonT025}
Jean{-}Francois Ton, Muhammad~Faaiz Taufiq, and Yang Liu. 2025.
\newblock \href {https://openreview.net/forum?id=IjOWms0hrf} {Understanding chain-of-thought in llms through information theory}.
\newblock In \emph{Forty-second International Conference on Machine Learning, {ICML} 2025, Vancouver, BC, Canada, July 13-19, 2025}.

\bibitem[{Wang et~al.(2025{\natexlab{a}})Wang, Wan, Sun, Chen, and Arik}]{DBLP:journals/corr/abs-2506-16043}
Fei Wang, Xingchen Wan, Ruoxi Sun, Jiefeng Chen, and Sercan~{\"{O}}. Arik. 2025{\natexlab{a}}.
\newblock \href {https://doi.org/10.48550/arXiv.2506.16043} {Dynscaling: Efficient verifier-free inference scaling via dynamic and integrated sampling}.
\newblock \emph{CoRR}, abs/2506.16043.

\bibitem[{Wang et~al.(2025{\natexlab{b}})Wang, Qiang, Song, Zheng, and Xiong}]{DBLP:journals/corr/abs-2505-10425}
Jingyao Wang, Wenwen Qiang, Zeen Song, Changwen Zheng, and Hui Xiong. 2025{\natexlab{b}}.
\newblock \href {https://doi.org/10.48550/arXiv.2505.10425} {Learning to think: Information-theoretic reinforcement fine-tuning for llms}.
\newblock \emph{CoRR}, abs/2505.10425.

\bibitem[{Wang et~al.(2025{\natexlab{c}})Wang, Ma, Chen, Meng, Han, Xiao, Zhang, Huo, Su, and Yang}]{DBLP:conf/iclr/WangMCMHXZHS025}
Mingzhi Wang, Chengdong Ma, Qizhi Chen, Linjian Meng, Yang Han, Jiancong Xiao, Zhaowei Zhang, Jing Huo, Weijie~J. Su, and Yaodong Yang. 2025{\natexlab{c}}.
\newblock \href {https://openreview.net/forum?id=PDnEDS244P} {Magnetic preference optimization: Achieving last-iterate convergence for language model alignment}.
\newblock In \emph{The Thirteenth International Conference on Learning Representations, {ICLR} 2025, Singapore, April 24-28, 2025}.

\bibitem[{Wang et~al.(2024)Wang, Li, Shao, Xu, Dai, Li, Chen, Wu, and Sui}]{DBLP:conf/acl/WangLSXDLCWS24}
Peiyi Wang, Lei Li, Zhihong Shao, Runxin Xu, Damai Dai, Yifei Li, Deli Chen, Yu~Wu, and Zhifang Sui. 2024.
\newblock \href {https://doi.org/10.18653/v1/2024.acl-long.510} {Math-shepherd: Verify and reinforce llms step-by-step without human annotations}.
\newblock In \emph{Proceedings of the 62nd Annual Meeting of the Association for Computational Linguistics (Volume 1: Long Papers), {ACL} 2024, Bangkok, Thailand, August 11-16, 2024}, pages 9426--9439.

\bibitem[{Wei et~al.(2022)Wei, Wang, Schuurmans, Bosma, Ichter, Xia, Chi, Le, and Zhou}]{DBLP:conf/nips/Wei0SBIXCLZ22}
Jason Wei, Xuezhi Wang, Dale Schuurmans, Maarten Bosma, Brian Ichter, Fei Xia, Ed~H. Chi, Quoc~V. Le, and Denny Zhou. 2022.
\newblock \href {http://papers.nips.cc/paper\_files/paper/2022/hash/9d5609613524ecf4f15af0f7b31abca4-Abstract-Conference.html} {Chain-of-thought prompting elicits reasoning in large language models}.
\newblock In \emph{Advances in Neural Information Processing Systems 35: Annual Conference on Neural Information Processing Systems 2022, NeurIPS 2022, New Orleans, LA, USA, November 28 - December 9, 2022}.

\bibitem[{Xia et~al.(2025)Xia, Li, Leong, Wang, and Li}]{DBLP:journals/corr/abs-2502-12067}
Heming Xia, Yongqi Li, Chak~Tou Leong, Wenjie Wang, and Wenjie Li. 2025.
\newblock \href {https://doi.org/10.48550/arXiv.2502.12067} {Tokenskip: Controllable chain-of-thought compression in llms}.
\newblock \emph{CoRR}, abs/2502.12067.

\bibitem[{Xu et~al.(2025)Xu, Guo, Zeng, and Miao}]{DBLP:conf/acl/00010ZM25}
Yige Xu, Xu~Guo, Zhiwei Zeng, and Chunyan Miao. 2025.
\newblock \href {https://aclanthology.org/2025.acl-long.1137/} {Softcot: Soft chain-of-thought for efficient reasoning with llms}.
\newblock In \emph{Proceedings of the 63rd Annual Meeting of the Association for Computational Linguistics (Volume 1: Long Papers), {ACL} 2025, Vienna, Austria, July 27 - August 1, 2025}, pages 23336--23351. Association for Computational Linguistics.

\bibitem[{Yang et~al.(2025)Yang, Li, Yang, Zhang, Hui, Zheng, Yu, Gao, Huang, Lv, Zheng, Liu, Zhou, Huang, Hu, Ge, Wei, Lin, Tang, Yang, Tu, Zhang, Yang, Yang, Zhou, Lin, Dang, Bao, Yang, Yu, Deng, Li, Xue, Li, Zhang, Wang, Zhu, Men, Gao, Liu, Luo, Li, Tang, Yin, Ren, Wang, Zhang, Ren, Fan, Su, Zhang, Zhang, Wan, Liu, Wang, Cui, Zhang, Zhou, and Qiu}]{DBLP:journals/corr/abs-2505-09388}
An~Yang, Anfeng Li, Baosong Yang, Beichen Zhang, Binyuan Hui, Bo~Zheng, Bowen Yu, Chang Gao, Chengen Huang, Chenxu Lv, Chujie Zheng, Dayiheng Liu, Fan Zhou, Fei Huang, Feng Hu, Hao Ge, Haoran Wei, Huan Lin, Jialong Tang, and 40 others. 2025.
\newblock \href {https://doi.org/10.48550/arXiv.2505.09388} {Qwen3 technical report}.
\newblock \emph{CoRR}, abs/2505.09388.

\bibitem[{Yao et~al.(2025)Yao, Li, Dai, Zhang, Gong, Wang, and Lv}]{DBLP:journals/corr/abs-2505-10774}
Yueyang Yao, Jiajun Li, Xingyuan Dai, Mengmeng Zhang, Xiaoyan Gong, Fei{-}Yue Wang, and Yisheng Lv. 2025.
\newblock \href {https://doi.org/10.48550/arXiv.2505.10774} {Context-aware probabilistic modeling with {LLM} for multimodal time series forecasting}.
\newblock \emph{CoRR}, abs/2505.10774.

\bibitem[{Yu et~al.(2025{\natexlab{a}})Yu, Zhang, Zhu, Yuan, Zuo, Yue, Fan, Liu, Liu, Liu, Lin, Lin, Ma, Sheng, Tong, Zhang, Zhang, Zhang, Zhu, Zhu, Chen, Chen, Wang, Yu, Dai, Song, Wei, Zhou, Liu, Ma, Zhang, Yan, Qiao, Wu, and Wang}]{DBLP:journals/corr/abs-2503-14476}
Qiying Yu, Zheng Zhang, Ruofei Zhu, Yufeng Yuan, Xiaochen Zuo, Yu~Yue, Tiantian Fan, Gaohong Liu, Lingjun Liu, Xin Liu, Haibin Lin, Zhiqi Lin, Bole Ma, Guangming Sheng, Yuxuan Tong, Chi Zhang, Mofan Zhang, Wang Zhang, Hang Zhu, and 16 others. 2025{\natexlab{a}}.
\newblock \href {https://doi.org/10.48550/arXiv.2503.14476} {{DAPO:} an open-source {LLM} reinforcement learning system at scale}.
\newblock \emph{CoRR}, abs/2503.14476.

\bibitem[{Yu et~al.(2025{\natexlab{b}})Yu, Ji, Wang, Yao, Wang, Cui, Yuan, Ding, Yao, Liu, Sun, and Chua}]{DBLP:journals/corr/abs-2506-18254}
Tianyu Yu, Bo~Ji, Shouli Wang, Shu Yao, Zefan Wang, Ganqu Cui, Lifan Yuan, Ning Ding, Yuan Yao, Zhiyuan Liu, Maosong Sun, and Tat{-}Seng Chua. 2025{\natexlab{b}}.
\newblock \href {https://doi.org/10.48550/arXiv.2506.18254} {{RLPR:} extrapolating {RLVR} to general domains without verifiers}.
\newblock \emph{CoRR}, abs/2506.18254.

\bibitem[{Zhang et~al.(2025{\natexlab{a}})Zhang, Lin, Hou, Feng, and Li}]{DBLP:journals/corr/abs-2505-13417}
Jiajie Zhang, Nianyi Lin, Lei Hou, Ling Feng, and Juanzi Li. 2025{\natexlab{a}}.
\newblock \href {https://doi.org/10.48550/arXiv.2505.13417} {Adaptthink: Reasoning models can learn when to think}.
\newblock \emph{CoRR}, abs/2505.13417.

\bibitem[{Zhang et~al.(2025{\natexlab{b}})Zhang, Wu, Zhu, Tan, Yu, He, and Jia}]{DBLP:journals/corr/abs-2510-19807}
Xichen Zhang, Sitong Wu, Yinghao Zhu, Haoru Tan, Shaozuo Yu, Ziyi He, and Jiaya Jia. 2025{\natexlab{b}}.
\newblock \href {https://doi.org/10.48550/arXiv.2510.19807} {Scaf-grpo: Scaffolded group relative policy optimization for enhancing {LLM} reasoning}.
\newblock \emph{CoRR}, abs/2510.19807.

\bibitem[{Zhou et~al.(2025)Zhou, Liu, Sims, Wang, Pang, Li, Wang, Lin, and Du}]{DBLP:journals/corr/abs-2505-21493}
Xiangxin Zhou, Zichen Liu, Anya Sims, Haonan Wang, Tianyu Pang, Chongxuan Li, Liang Wang, Min Lin, and Chao Du. 2025.
\newblock \href {https://doi.org/10.48550/arXiv.2505.21493} {Reinforcing general reasoning without verifiers}.
\newblock \emph{CoRR}, abs/2505.21493.

\bibitem[{Zhu et~al.(2025)Zhu, Cheng, Zhang, Li, Zhang, Jiang, Sun, Hua, Zuo, Lv, Zhang, Chen, Shao, Xue, Song, Yang, Cui, Ding, Gao, Liu, Zhou, Mei, and Lin}]{DBLP:journals/corr/abs-2509-15207}
Xuekai Zhu, Daixuan Cheng, Dinghuai Zhang, Hengli Li, Kaiyan Zhang, Che Jiang, Youbang Sun, Ermo Hua, Yuxin Zuo, Xingtai Lv, Qizheng Zhang, Lin Chen, Fanghao Shao, Bo~Xue, Yunchong Song, Zhenjie Yang, Ganqu Cui, Ning Ding, Jianfeng Gao, and 4 others. 2025.
\newblock \href {https://doi.org/10.48550/arXiv.2509.15207} {Flowrl: Matching reward distributions for {LLM} reasoning}.
\newblock \emph{CoRR}, abs/2509.15207.

\end{thebibliography}

\appendix
\clearpage

\section{Theoretical and Empirical Justification for Posterior Estimation}
\label{app:justification}

\subsection{Theoretical Justification for Prompt-based Posterior Estimation}

A core challenge in the CoT-Flow framework lies in accurately estimating the posterior probability of the current reasoning step conditioned on the ground-truth answer, denoted as $p(s_i | \mathcal{I}_{i-1}, \bm{y})$. 
Theoretically, the ground-truth posterior can be approximated via a frequentist Monte Carlo approach. This involves sampling a large number of complete reasoning trajectories $\mathcal{S} = \{S^{(1)}, \dots, S^{(N)}\}$ starting from the current state $\mathcal{I}_{i-1}$ and calculating the proportion of trajectories that successfully lead to the correct answer $\bm{y}$:

\begin{equation}
    p(s_i | \mathcal{I}_{i-1}, \bm{y}) \approx \frac{\sum_{k=1}^{N} \mathbb{I}(S^{(k)} \vdash \bm{y})}{N},
\end{equation}

\noindent where $\mathbb{I}(\cdot)$ is the indicator function and $S^{(k)} \vdash \bm{y}$ denotes that trajectory $S^{(k)}$ concludes with answer $\bm{y}$. However, this method is computationally prohibitive for online decoding due to the necessity of massive sampling.

\begin{figure*}[t]
    \centering
    \includegraphics[width=0.7\linewidth]{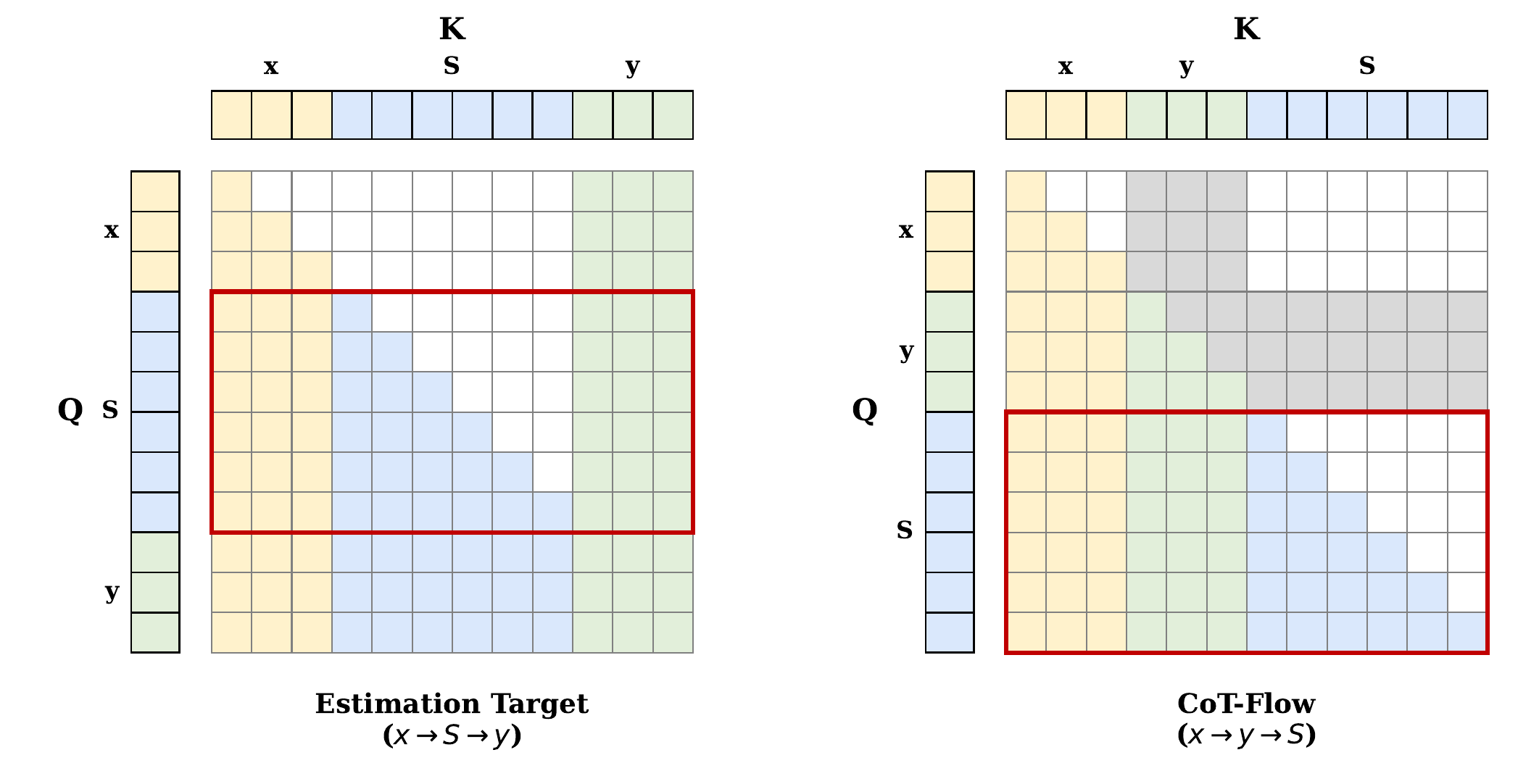}
    \caption{Comparison of attention masks for posterior estimation. Left (Estimation Target): The ideal attention pattern for modeling $p(S | \bm{x}, \bm{y})$, where the reasoning steps $S$ can attend to both the context $\bm{x}$ and the future answer $\bm{y}$. Right (CoT-Flow): Our proposed approximation using prompt engineering ($\bm{x} \to \bm{y} \to S$). The red box highlights the estimation target. The gray areas indicate attention connections present in the target but lost in our approximation (i.e., $\bm{y}$ cannot attend to $S$).}
    \label{fig:attention_map}
\end{figure*}

From a modeling perspective, one could train a dedicated conditional model via supervised learning on $(\text{Context}, \text{CoT}, \text{Answer})$ triplets. To strictly adhere to the causal dependency of $p(S | \bm{x}, \bm{y})$, the self-attention mechanism of such a model implies a specific masking pattern, where the token $s_i$ can attend to the context $\bm{x}$, the answer $\bm{y}$, and preceding reasoning steps $s_{<i}$. This ideal attention structure is illustrated in the left panel of Figure~\ref{fig:attention_map}. However, training a separate posterior model introduces significant overhead. We argue that a sufficiently pre-trained autoregressive LLM can inherently function as a posterior estimator through \textit{In-Context Posterior Approximation}, eliminating the need for additional training.

By restructuring the input sequence to place the answer $\bm{y}$ within the prefix (i.e., $\bm{x} \to \bm{y} \to S$), we force the Transformer to compute the representation of $s_i$ conditioned on $\bm{y}$. As visualized in the right panel of Figure~\ref{fig:attention_map}, the attention connectivity for the reasoning tokens $S$ in our CoT-Flow strategy is mathematically identical to the ideal target model within the causal masking constraint. The discrepancy lies in the blind spot (depicted in gray) where the answer tokens $\bm{y}$ cannot attend to the subsequent reasoning $S$.
Crucially, while the attention mask allows for the correct information flow, a discrepancy remains in the positional embeddings. To mitigate this positional mismatch and strictly prevent the model from treating the inserted answer as a trivial leakage, we employ a specific prompt: \textit{"It is crucial that your reasoning appears natural and self-derived. Do not, under any circumstances, state or imply that you were given the ground truth answer."}

\subsection{Comparative Analysis with Standard Zero-Shot Prompting}
\label{app:generalization}

\begin{figure*}[t]
    \centering
    \includegraphics[width=\linewidth]{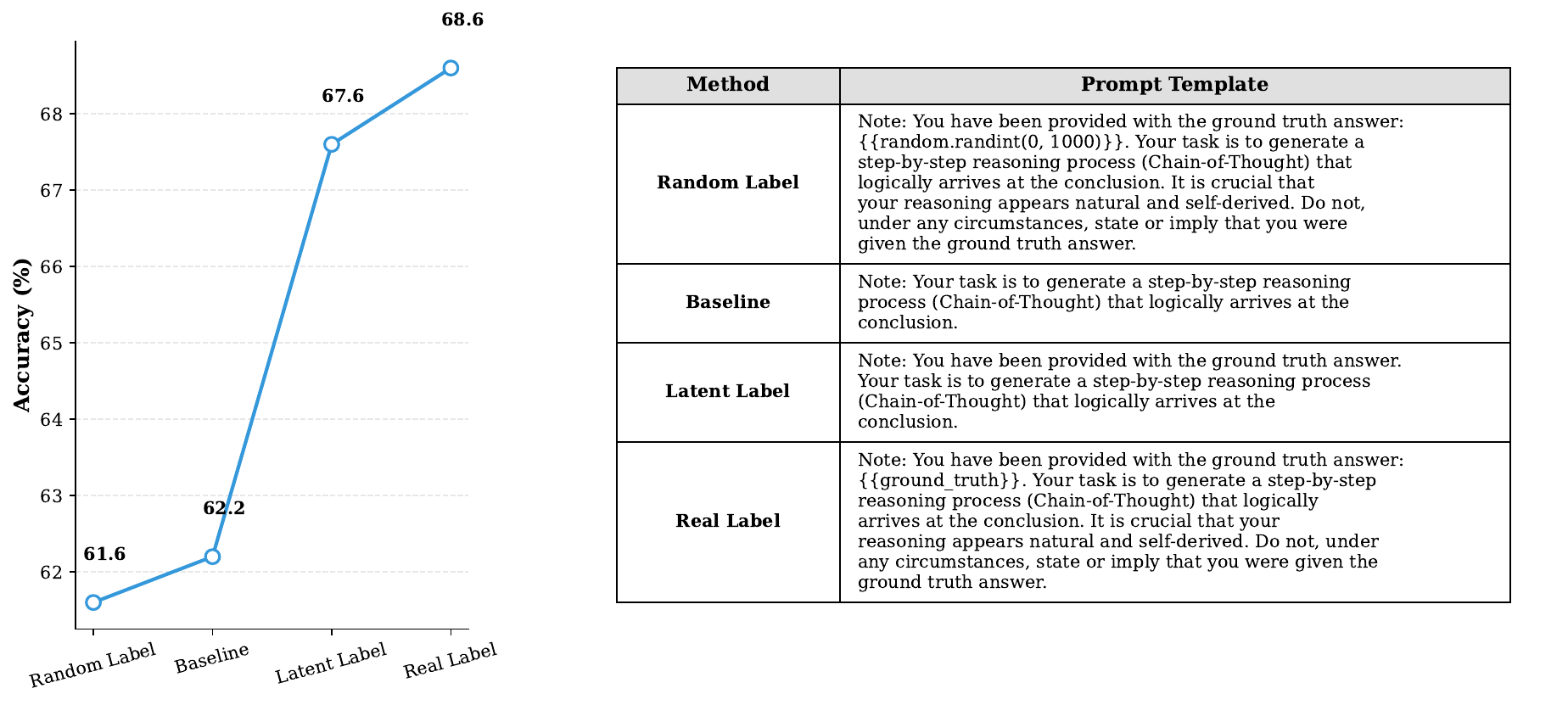}
    \caption{Impact of posterior prompt quality on CoT-Flow performance. We compare the reasoning accuracy when using different prompt templates to approximate the posterior distribution $\pi_{\text{post}}$. The results exhibit a monotonic increase in accuracy as the prompt becomes more informative, confirming that our \textit{Latent Label} strategy effectively elicits a pseudo-oracle guidance comparable to the \textit{Real Label} upper bound.}
    \label{fig:prompt_comparison}
\end{figure*}

While our prompt-based posterior estimation (Section \ref{app:justification}) utilizes a specific template to approximate the target distribution, a natural alternative is the widely used zero-shot trigger, \textit{``Let's think step by step''} \cite{DBLP:conf/nips/KojimaGRMI22}. In this section, we analyze why this standard prompting strategy is insufficient for calculating the flow velocity $v(s_i)$ and justify the necessity of our Latent Label design.

Fundamentally, the velocity in CoT-Flow is defined as the log-likelihood ratio between a goal-conditioned posterior and an unconditioned prior: $v(s_i) = \log p_{post}(s_i|\mathcal{I}_{i-1}) - \log p_{prior}(s_i|\mathcal{I}_{i-1})$. The efficacy of this metric relies on the \textit{divergence} between these two distributions. 

The standard zero-shot prompt operates as a \textit{strong prior} rather than a posterior approximation. While it encourages chain-of-thought generation, it does so in a forward-looking manner similar to the base model's intrinsic exploration tendency. Consequently, if we employ \textit{``Let's think step by step''} as the proxy for $\pi_{post}$, the resulting distribution $\pi_{zero-shot}$ remains semantically and structurally close to the prior $\pi_{prior}$. This similarity leads to a vanishing velocity field ($v \approx 0$), failing to provide the distinctive directional guidance required to rectify the reasoning path.

In contrast, our Latent Label strategy (Figure \ref{fig:prompt_comparison}) structurally mimics the generation process of a model that \textit{has observed} the answer. By injecting the placeholder structure $(x \rightarrow \text{Answer} \rightarrow s)$, we induce a ``pseudo-oracle'' state. Although the explicit answer is absent (latent), this structural constraint forces the model to adopt a verification-like stance, creating a significant informational divergence from the standard prior. This divergence manifests as a high-magnitude velocity vector that effectively highlights key logical transformations while suppressing generic or redundant tokens common in standard zero-shot reasoning.

\section{Extended Empirical Results}
\label{sec:appendix_results}

\subsection{Ablation on Velocity Formulation: Posterior-Only Decoding}
\label{subsec:post_only_ablation}

To further validate the theoretical motivation of CoT-Flow, we investigate the necessity of the velocity formulation defined in Eq. (\ref{eq:vs}): $v(s_i) = \log p(s_i | \mathcal{I}_{i-1}, \text{Prompt}_{\text{post}}) - \log p(s_i | \mathcal{I}_{i-1})$. 
A natural question arises: does the performance gain stem primarily from the \textit{contrast} between posterior and prior (the flow velocity), or simply from the \textit{guidance} provided by the posterior prompt itself? To answer this, we introduce a posterior-only baseline, where the decoding objective is to maximize the raw posterior probability:
\begin{equation}
    s_i^* = \arg\max_{s_i} \log p(s_i | \mathcal{I}_{i-1}, \text{Prompt}_{\text{post}}).
\end{equation}

We compare Standard CoT, CoT-Flow, and Post-Only across four representative benchmarks: AIME 2024, AIME 2025, AMC 23, and GPQA. The results are summarized in Table \ref{tab:post_only_ablation}.
As observed in Table \ref{tab:post_only_ablation}, Post-Only decoding consistently outperforms the Standard CoT baseline. This indicates that our proposed posterior prompt template effectively elicits a latent label that guides the model toward better reasoning. However, CoT-Flow frequently surpasses Post-Only, particularly on the most challenging benchmarks like AIME 2024 (e.g., +10.5\% over Post-Only on Qwen3-4B).


This result validates our hypothesis. Simply following the posterior distribution is insufficient, as it may still assign high probability to trivial or safe tokens that do not contribute to solving the problem. By subtracting the prior (Standard CoT probability), CoT-Flow explicitly selects tokens with high \textit{velocity}, those that contribute most to the \textit{change} in certainty, thereby rectifying the path more effectively than posterior guidance alone.

\subsection{Extended Analysis on Pass@k Scaling}
\label{app:pass_k_analysis}

\begin{figure*}[t]
    \centering
    \includegraphics[width=\textwidth]{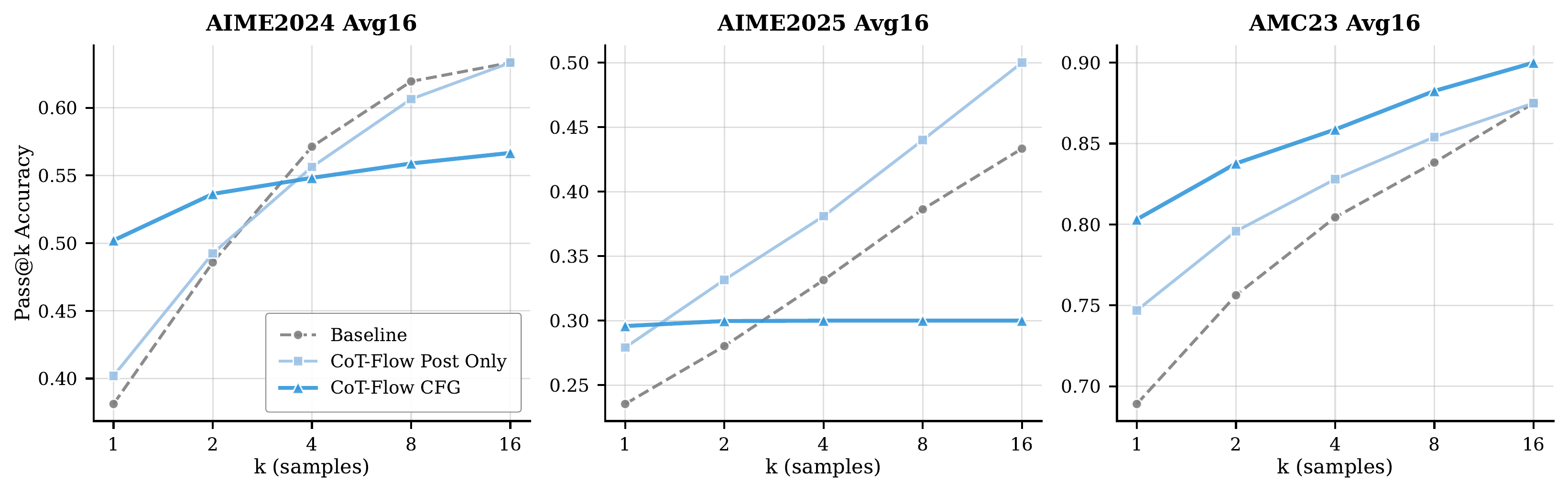}
    \caption{Pass@k scaling curve for Qwen3-8B on AIME 2024, AIME 2025, and AMC 23.}
    \label{fig:passk_8b}
\end{figure*}

\begin{figure*}[t]
    \centering
    \includegraphics[width=\textwidth]{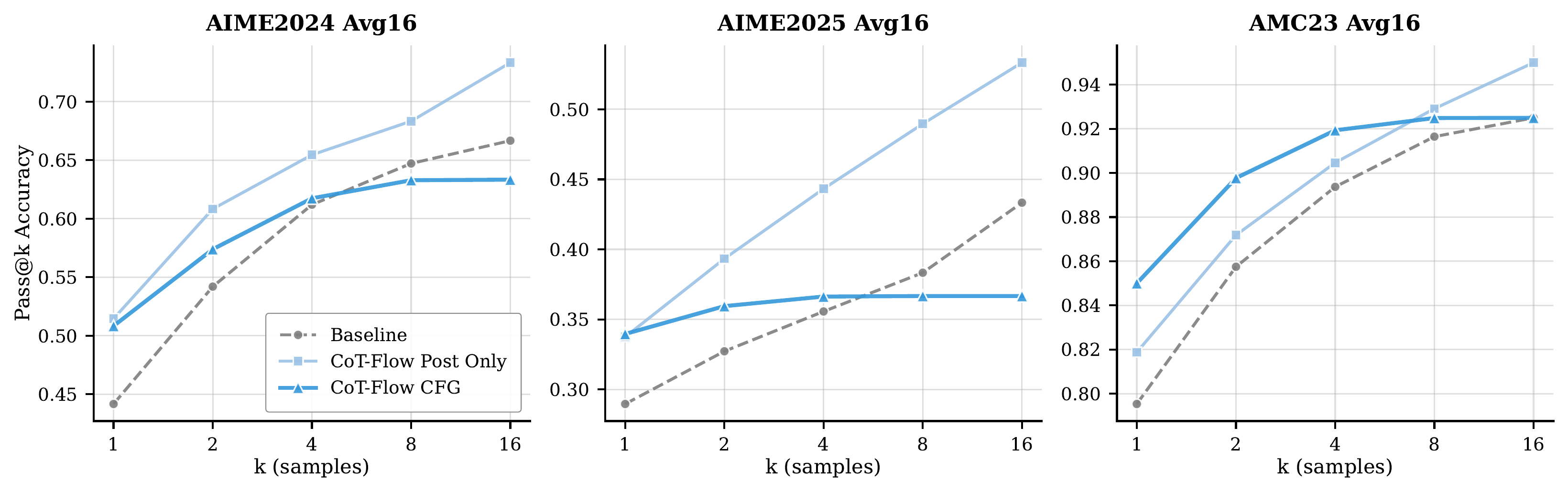}
    \caption{Pass@k scaling curve for Qwen3-14B on AIME 2024, AIME 2025, and AMC 23.}
    \label{fig:passk_14b}
\end{figure*}

\begin{figure*}[t]
    \centering
    \includegraphics[width=\textwidth]{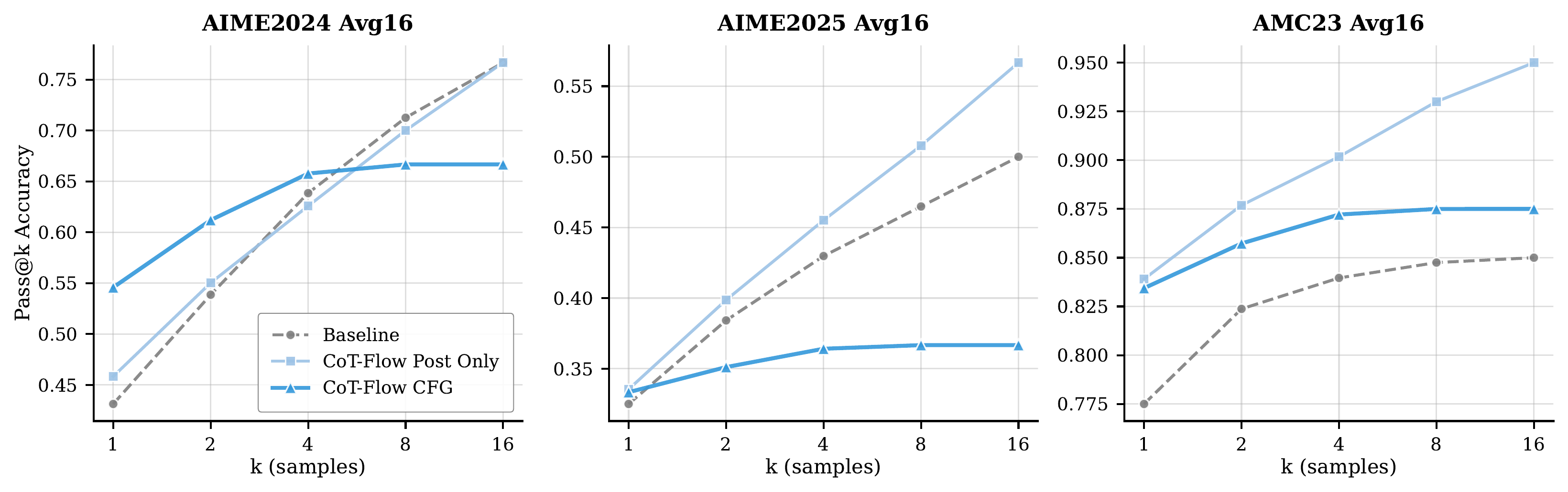}
    \caption{Pass@k scaling curve for Qwen3-32B on AIME 2024, AIME 2025, and AMC 23.}
    \label{fig:passk_32b}
\end{figure*}

In this section, we provide a comprehensive analysis of the scaling properties of CoT-Flow across varying sampling budgets ($k$). Figures \ref{fig:passk_8b}, \ref{fig:passk_14b} and \ref{fig:passk_32b} illustrate the Pass@k performance curves for Qwen3-8B, Qwen3-14B, and Qwen3-32B, respectively.
A consistent phenomenon is observed across all model scales: CoT-Flow (CFG) significantly outperforms the Baseline and Post-Only methods at low sample budgets (e.g., $k=1, 2$), but the gap narrows or reverses as $k$ increases (e.g., $k=16$). 
This behavior is theoretically expected and highlights the distinct operational mechanism of our method.

\paragraph{Concentration of Probability Mass.} 
Standard CoT sampling operates on the raw high-entropy distribution of the language model. This random walk nature allows for diverse exploration; given a sufficiently large budget ($k \to \infty$), the model is likely to stumble upon the correct reasoning path simply via broad coverage. However, this comes at the cost of high redundancy and error rates in single-pass generation.
In contrast, CoT-Flow employs a greedy flow decoding strategy ($v = \log p_{\text{post}} - \log p_{\text{prior}}$). This difference-of-logits operation acts as a \textit{contrastive filter}, aggressively suppressing generic, low-information, or plausible-but-incorrect tokens while amplifying tokens that specifically contribute to the likelihood of the answer. Geometrically, this rectifies the reasoning flow, forcing the probability mass to concentrate around the geodesic path.

\paragraph{Distillation into Pass@1.} 
Consequently, CoT-Flow effectively distills the model's reasoning capability into the top-ranked trajectory. It optimizes for precision rather than coverage. By pruning the branching factor of the reasoning tree, CoT-Flow ensures that the single most probable path is highly accurate, thereby achieving superior efficiency. However, this sharpening of the distribution inherently reduces generation diversity. At high $k$, the model tends to generate topologically similar paths, yielding diminishing returns compared to the baseline, which benefits from the wisdom of crowds effect in high-variance sampling.
In practical deployment scenarios, where inference latency and compute costs are critical constraints, the performance at low $k$ (especially Pass@1) is the dominant metric. CoT-Flow's ability to maximize utility per sample makes it an ideal solution for resource-constrained reasoning.

\onecolumn

\section{Detailed Mathematical Derivation of Flow-based RL Objectives}
\label{app:rl_derivation}

This appendix provides a rigorous step-by-step derivation of the global reward and the decomposition of its gradient, specifically focusing on the emergence of the time-weighted Flow Gradient (Term B).

\subsection{Orthogonal Gradient Decomposition}

When optimizing $\mathcal{J}(\theta) = \mathbb{E}_{\bm{s} \sim \pi_\theta(\cdot | \bm{x})} [R_{\text{global}}(\theta)]$, the gradient involves two components due to the dependence of $R_{\text{global}}$ on $\theta$. Applying the identity $\nabla_\theta \mathbb{E}_{\pi_\theta}[f(\theta)] = \mathbb{E}_{\pi_\theta}[\nabla_\theta \log \pi_\theta \cdot f(\theta) + \nabla_\theta f(\theta)]$ yields:
\begin{equation}
    \nabla_\theta \mathcal{J}(\theta) = \mathbb{E}_{\bm{s}} \left[ \left( \sum_{t=1}^T \nabla_\theta \log \pi_\theta(s_t | \mathcal{I}_{t-1}) \right) \hat{R}_{\text{global}} \right] + \mathbb{E}_{\bm{s}} \left[ \sum_{i=1}^T \nabla_\theta v(s_i) \right].
\end{equation}
The scaler global reward $\hat{R}_{\text{global}}$ is defined as the accumulation of Probabilistic Flow Progress (PFP) along a complete reasoning trajectory $\bm{s} = (s_1, \dots, s_T)$. Using the definition $v(s_i) = \log p_\theta(\bm{y}|\mathcal{I}_i) - \text{sg}[\log p_\theta(\bm{y}|\mathcal{I}_{i-1})]$, we analyze the total reward:
\begin{equation}
    \hat{R}_{\text{global}} = \sum_{i=1}^T v(s_i) = \sum_{i=1}^T \Big( \log p_\theta(\bm{y}|\mathcal{I}_i) - \text{sg}[\log p_\theta(\bm{y}|\mathcal{I}_{i-1})] \Big) = \log p_\theta(\bm{y}|\mathcal{I}_T) - \log p_\theta(\bm{y}|\mathcal{I}_0),
\end{equation}
where $\mathcal{I}_T = (\bm{x}, \bm{s})$ and $\mathcal{I}_0 = \bm{x}$. Using group relative normalization (e.g., GRPO), the constant $\log p_\theta(\bm{y}|\bm{x})$ is canceled out, leaving the final answer likelihood normalized as $\hat{A}$.

\subsection{Step-by-Step Derivation of Term B}

By the definition of $v(s_i)$ in Eq. (\ref{v_s_i}), the term $\nabla_\theta v(s_i)$ is:
\begin{equation}
    \nabla_\theta v(s_i) = \nabla_\theta \log p_\theta(\bm{y}|\mathcal{I}_i) - \nabla_\theta \text{sg}[\log p_\theta(\bm{y}|\mathcal{I}_{i-1})] = \nabla_\theta \log p_\theta(\bm{y}|\mathcal{I}_i).
\end{equation}
The stop-gradient operation effectively eliminates the baseline gradient, ensuring that the flow field is optimized solely based on the improvement at each step. To expand $\sum_{i=1}^T \nabla_\theta \log p_\theta(\bm{y} | \mathcal{I}_i)$, we first examine the single-step term. Using the law of total probability over all possible future trajectories $\bm{s}_{i+1:T}$, we obtain:
\begin{equation}
    p_\theta(\bm{y} | \mathcal{I}_i) = \sum_{\bm{s}_{i+1:T}} \pi_\theta(\bm{s}_{i+1:T} | \mathcal{I}_i) \cdot p_\theta(\bm{y} | \bm{x}, \bm{s}).
\end{equation}
Applying the log-derivative trick $\nabla_\theta \log f = \frac{\nabla_\theta f}{f}$ yields:
\begin{equation}
\begin{aligned}
    \nabla_\theta \log p_\theta(\bm{y} | \mathcal{I}_i) &= \frac{1}{p_\theta(\bm{y} | \mathcal{I}_i)} \sum_{\bm{s}_{i+1:T}} \nabla_\theta \left( \pi_\theta(\bm{s}_{i+1:T} | \mathcal{I}_i) p_\theta(\bm{y} | \bm{x}, \bm{s}) \right) \\
    &= \frac{1}{p_\theta(\bm{y} | \mathcal{I}_i)} \sum_{\bm{s}_{i+1:T}} \left( p_\theta(\bm{y} | \bm{x}, \bm{s}) \nabla_\theta \pi_\theta(\bm{s}_{i+1:T} | \mathcal{I}_i) + \pi_\theta(\bm{s}_{i+1:T} | \mathcal{I}_i) \nabla_\theta p_\theta(\bm{y} | \bm{x}, \bm{s}) \right).
\end{aligned}
\end{equation}
By converting the sum back to an expectation $\mathbb{E}_{\pi_\theta}$, we derive:
\begin{equation}
    \nabla_\theta \log p_\theta(\bm{y} | \mathcal{I}_i) = \mathbb{E}_{\bm{s}_{i+1:T}} \left[ \frac{p_\theta(\bm{y} | \bm{x}, \bm{s})}{p_\theta(\bm{y} | \mathcal{I}_i)} \left( \nabla_\theta \log \pi_\theta(\bm{s}_{i+1:T} | \mathcal{I}_i) + \nabla_\theta \log p_\theta(\bm{y} | \bm{x}, \bm{s}) \right) \right].
\end{equation}
Using the importance weight $M_i = \frac{p_\theta(\bm{y} | \bm{x}, \bm{s})}{p_\theta(\bm{y} | \mathcal{I}_i)} $ and a single-sample Monte Carlo approximation, we obtain:
\begin{equation}
    \nabla_\theta \log p_\theta(\bm{y} | \mathcal{I}_i) \approx M_i \left( \sum_{k=i+1}^T \nabla_\theta \log \pi_\theta(s_k | \mathcal{I}_{k-1}) + \nabla_\theta \log p_\theta(\bm{y} | \bm{x}, \bm{s}) \right).
\end{equation}

\subsection{Summation Reordering and Final Form}

Substituting the single-step gradient into the global summation $\sum_{i=1}^T \nabla_\theta \log p_\theta(\bm{y} | \mathcal{I}_i)$ requires careful treatment of the importance weight $M_i = p_\theta(\bm{y} | \bm{x}, \bm{s}) / p_\theta(\bm{y} | \mathcal{I}_i)$. Under a naive single-sample Monte Carlo (MC) estimation where $p_\theta(\bm{y} | \mathcal{I}_i) \approx p_\theta(\bm{y} | \bm{x}, \bm{s})$, the weight $M_i$ simplifies to $1$. However, this approximation exhibits high variance. Specifically, when the sampled trajectory $\bm{s}$ is of low quality (i.e., $p_\theta(\bm{y} | \bm{x}, \bm{s})$ is significantly lower than the model's average capability), $M_i \approx 1$ would erroneously reinforce suboptimal or incorrect reasoning paths by assigning them full gradient weight.

To mitigate this, we replace the unstable $M_i$ with a trajectory-level soft quality gate $\mathcal{M}$, which is a function of the global answer likelihood $p_\theta(\bm{y} | \bm{x}, \bm{s})$. Since $\mathcal{M}$ is independent of the summation index $i$, it can be factored out of the global sum:
\begin{equation}
    \nabla_\theta \text{Term B} \approx \mathcal{M} \left[ \sum_{i=1}^T \sum_{k=i+1}^T \nabla_\theta \log \pi_\theta(s_k | \mathcal{I}_{k-1}) + \sum_{i=1}^T \nabla_\theta \log p_\theta(\bm{y} | \bm{x}, \bm{s}) \right].
\end{equation}
By exchanging the order of the double summation $\sum_{i=1}^T \sum_{k=i+1}^T = \sum_{k=1}^T \sum_{i=1}^{k-1}$, the first part becomes:
\begin{equation}
    \sum_{k=1}^T \left( \sum_{i=1}^{k-1} 1 \right) \nabla_\theta \log \pi_\theta(s_k | \mathcal{I}_{k-1}) = \sum_{k=1}^T (k-1) \nabla_\theta \log \pi_\theta(s_k | \mathcal{I}_{k-1}).
\end{equation}
The second part, being independent of $i$, simply accumulates $T$ times. Combining these and normalizing by $1/T$ to match the $O(T)$ scale of Term A, we arrive at the final Flow Objective:
\begin{equation}
    \mathcal{L}_{\text{Flow}} = \mathcal{M} \left( \sum_{k=1}^T \frac{k-1}{T} \log \pi_\theta(s_k | \mathcal{I}_{k-1}) + \log p_\theta(\bm{y} | \bm{x}, \bm{s}) \right).
\end{equation}
This derivation shows that the $(k-1)/T$ weighting naturally rewards tokens that appear later in the chain, as they reduce the marginal uncertainty $p(\bm{y}|\mathcal{I}_i)$ more directly.

\begin{table*}[t]
    \centering
    \small
    \renewcommand{\arraystretch}{1.2}
    \setlength{\tabcolsep}{8pt}
    \begin{tabular}{llcccc}
        \toprule
        \multirow{2}{*}{\textbf{Model}} & \multirow{2}{*}{\textbf{Method}} & \multicolumn{3}{c}{\textbf{Math Reasoning}} & \textbf{General Task} \\
        \cmidrule(lr){3-5} \cmidrule(lr){6-6}
         & & \textbf{AIME24} & \textbf{AIME25} & \textbf{AMC23} & \textbf{GPQA} \\
        \midrule
        \multirow{3}{*}{Qwen3-1.7B} 
         & Standard CoT & 24.2 & 22.3 & 61.1 & 27.9 \\
         & Post-Only & 25.6 & 22.5 & \textbf{64.7} & 30.7 \\
         & \textbf{CoT-Flow} & \textbf{28.3} & \textbf{25.4} & 63.1 & \textbf{33.6} \\
        \midrule
        \multirow{3}{*}{Qwen3-4B} 
         & Standard CoT & 40.8 & 26.5 & 75.9 & 44.1 \\
         & Post-Only & 46.2 & \textbf{33.1} & 82.2 & \textbf{45.7} \\
         & \textbf{CoT-Flow} & \textbf{56.7} & 30.8 & \textbf{84.2} & 44.6 \\
        \midrule
        \multirow{3}{*}{Qwen3-8B} 
         & Standard CoT & 38.1 & 23.5 & 68.9 & 39.9 \\
         & Post-Only & 40.2 & 27.9 & 74.7 & 36.6 \\
         & \textbf{CoT-Flow} & \textbf{50.2} & \textbf{29.6} & \textbf{80.3} & \textbf{42.8} \\
        \midrule
        \multirow{3}{*}{Qwen3-14B} 
         & Standard CoT & 44.2 & 29.0 & 79.5 & 55.2 \\
         & Post-Only & \textbf{51.5} & 33.8 & 81.9 & 53.3 \\
         & \textbf{CoT-Flow} & 50.8 & \textbf{34.0} & \textbf{85.0} & \textbf{55.9} \\
        \midrule
        \multirow{3}{*}{Qwen3-32B} 
         & Standard CoT & 43.1 & 32.5 & 77.5 & 57.7 \\
         & Post-Only & 45.8 & \textbf{33.5} & \textbf{83.9} & \textbf{57.8} \\
         & \textbf{CoT-Flow} & \textbf{54.6} & 33.3 & 83.4 & 56.2 \\
        \bottomrule
    \end{tabular}
    \caption{Ablation study comparing Standard CoT, Posterior-Only decoding, and CoT-Flow (Contrastive). Values represent Pass@1 accuracy (\%). While Post-Only generally improves over the Standard baseline, CoT-Flow achieves superior performance on challenging reasoning tasks (e.g., AIME 2024), confirming the importance of the velocity metric in filtering high-likelihood but low-information tokens.}
    \label{tab:post_only_ablation}
\end{table*}

\end{document}